\documentclass{IEEEtran}
\usepackage{cite}
\usepackage{amsmath}
\usepackage{amssymb}
\usepackage{graphicx}
\usepackage{booktabs}
\usepackage[table]{xcolor}
\usepackage{xcolor}
\usepackage{amsmath}
\usepackage{multirow}
\usepackage{booktabs}
\usepackage{multirow}
\usepackage{pifont} 
\newcommand{\cmark}{\ding{51}} 
\newcommand{\xmark}{\ding{55}} 
\usepackage{pifont}
\usepackage[colorlinks=true, linkcolor=blue, citecolor=blue, urlcolor=blue]{hyperref}

\begin{document}

\title{WS-Net: Weak-Signal Representation Learning and Gated Abundance Reconstruction for Hyperspectral Unmixing via State-Space and Weak Signal Attention Fusion}

\author{
    Zekun Long,~\IEEEmembership{Student Member,~IEEE}, 
    Ali Zia, 
    Guanyiman Fu,
    Vivien Rolland,
    Jun Zhou,~\IEEEmembership{Fellow,~IEEE}
    
    \thanks{Z. Long, G. Fu, and J. Zhou are with the School of Information and Communication Technology, Griffith University, Brisbane, QLD 4111, Australia 
    (e-mail: zekun.long@griffithuni.edu.au; guanyiman.fu@griffith.edu.au; jun.zhou@griffith.edu.au).}
    
    \thanks{A. Zia is with the School of Computing, Engineering and Mathematical Sciences, La Trobe University, Melbourne, VIC 3086, Australia 
    (e-mail: a.zia@latrobe.edu.au).}

    \thanks{V. Rolland is with CSIRO Agriculture and Food, Canberra, ACT 2601, Australia 
    (e-mail: Vivien.Rolland@csiro.au).}
}
\maketitle
\begin{abstract}
 Weak spectral responses in hyperspectral images are often obscured by dominant endmembers and sensor noise, resulting in inaccurate abundance estimation. This paper introduces WS-Net, a deep unmixing framework specifically designed to address weak-signal collapse through state-space modelling and Weak Signal Attention fusion. The network features a multi-resolution wavelet-fused encoder that captures both high-frequency discontinuities and smooth spectral variations with a hybrid backbone that integrates a Mamba state-space branch for efficient long-range dependency modelling. It also incorporates a Weak Signal Attention branch that selectively enhances low-similarity spectral cues. A learnable gating mechanism adaptively fuses both representations, while the decoder leverages KL-divergence-based regularisation to enforce separability between dominant and weak endmembers. Experiments on one simulated and two real datasets (synthetic dataset, Samson, and Apex) demonstrate consistent improvements over six state-of-the-art baselines, achieving up to 55\% and 63\% reductions in RMSE and SAD, respectively. The framework maintains stable accuracy under low-SNR conditions, particularly for weak endmembers, establishing WS-Net as a robust and computationally efficient benchmark for weak-signal hyperspectral unmixing. The source code and trained models will be made publicly available at \url{https://github.com/leolala/WS-Mamba.git}


\end{abstract}

\section{Introduction}
\label{sec:intro}

Hyperspectral imaging (HSI) has emerged as a powerful remote sensing technique, offering rich spectral and spatial information that enables detailed analysis of Earth's surface~\cite{jia2020status, goetz2009three}. By capturing hundreds of contiguous spectral bands per pixel, HSI facilitates diverse applications including land use monitoring~\cite{lv2023spatialhsilanduse}, mineral detection~\cite{bai2023miningmapping}, vegetation mapping~\cite{roberts1998mapping}, and climate change assessment~\cite{yang2013role}. However, remote sensing hyperspectral images often have a large instantaneous field-of-view and suffer from low spatial resolution~\cite{stuckens2000integrating}, resulting in a single pixel often containing a mixture of different land cover types and materials~\cite{bhatt2020review}~\cite{Bioucashu2012}, which complicates detailed analysis.

A key remedy is \textbf{hyperspectral unmixing} (HU), which decomposes each pixel’s reflectance spectrum into a set of constituent material spectra (endmembers)~\cite{miao2007endmember} and their corresponding fractional abundances~\cite{borsoi2021spectral}. This process underpins accurate material identification and quantification in complex scenes and supports downstream tasks such as classification~\cite{bioucas2013hyperspectral}, detection~\cite{manolakis2002detection}, and anomaly analysis~\cite{qu2018hyperspectral,liu2024uadnet}. In remote sensing applications, the widely adopted \textbf{linear mixing model} (LMM) assumes that each observed pixel is a convex combination of pure material spectra~\cite{keshava2002spectral}. When endmembers are known, the abundances can be estimated by minimizing the least-squares discrepancy between the measured spectrum and its linear reconstruction. To ensure physical interpretability, two constraints are typically imposed: the abundance non-negativity constraint (ANC)~\cite{keshava2002spectral}, which prohibits negative abundances, and the abundance sum-to-one constraint (ASC)~\cite{chang2010review}, which enforces that endmember contributions fully account for the observed spectrum. Traditional unmixing algorithms commonly incorporate ANC and ASC and employ various matrix-factorization~\cite{jia2008constrained} or constrained-optimisation strategies to determine abundance~\cite{feng2022hyperspectral}.

Despite their success, these models face significant limitations when dealing with low-reflectance materials, whose spectral signatures are inherently weak and frequently submerged within noise or overwhelmed by stronger endmembers~\cite{wen2018fast,funk2002clustering}. Such components are especially common in real-world environments; examples include trace-level pollutants, shaded water bodies, dark minerals, or low-abundance anthropogenic materials~\cite{zaresparsitylow}. Their subtle spectral contributions are easily masked in mixed pixels, making them difficult to detect and often leading to underestimation or exclusion during unmixing.

Recent deep learning-based approaches have made progress in addressing the non-linearities and spatial dependencies inherent in HSI data~\cite{guo2015hyperspectral, ozkan2018endnet, palsson2020convolutional, rasti2022misicnet}. 
However, most existing models still assume that dominant, high reflectance signatures are the primary contributors to pixel composition. 
CNN-based methods often compress spectral dimensions prematurely or lack mechanisms for selectively enhancing weak signals, resulting in degraded performance under low signal-to-noise ratio (SNR) conditions~\cite{palsson2020convolutional}. Moreover, Transformer-based methods such as DeepTrans-HSU~\cite {9848995} and Entropy-Driven Adaptive Algorithms (EDAA)~\cite{zouaoui2023entropic} introduce global attention mechanisms to improve long-range feature modelling, but these approaches often come at the cost of computational complexity and may overlook weak spectral cues.

 While the linear mixing model (LMM) provides a useful first-order approximation of pixel composition, it neglects nonlinear scattering and illumination effects that become dominant when endmembers exhibit low reflectance or low fractional abundance. These nonlinear interactions distort the linear convex geometry assumed in conventional formulations, making the estimation of abundances under weak-signal regimes inherently unstable. In this work, we retain the interpretability of the LMM but introduce a nonlinear corrective mapping, parameterized by state-space and Weak Signal Attention operators, to regularize the inversion and recover physically consistent abundances.

In particular, low-reflectance endmembers present two core challenges for unmixing: (1) \textit{spectral weakness}, which increases sensitivity to noise and limits detectability, (2) \textit{spatial sparsity}, as such materials often occupy small, scattered regions across the scene. These issues collectively contribute to what we term the ``weak signal collapse'' problem in hyperspectral unmixing. It represents a failure mode wherein low-magnitude endmembers are either underrepresented or completely missed in the unmixing process due to interference from dominant spectral signatures. In this work, we \emph{formally define a weak-signal hyperspectral unmixing scenario} as one in which at least one endmember exhibits (i) an average reflectance magnitude below 0.1 across more than 70 \% of spectral bands or (ii) contributes less than 10 \% fractional abundance in over 60 \% of pixels. These low-SNR, low-abundance conditions often lead to endmember under-representation. Addressing this regime requires architectures explicitly designed to preserve and reconstruct low-energy spectral components.

Despite growing interest in deep learning-based weak hyperspectral unmixing, several critical limitations remain unresolved. Most encoder designs do not explicitly enhance weak signals or suppress high-frequency spectral noise, which can result in the loss of crucial information early in the processing pipeline. In addition, attention-based methods, though capable of modelling long-range dependencies, tend to be computationally expensive and may fail to capture spatially sparse or fragmented patterns effectively. Decoder architectures also typically treat all spectral components equally, without explicitly distinguishing between dominant and weak endmembers, which leads to the underestimation of low-reflectance materials. Furthermore, many existing methods lack rigorous evaluation under low signal-to-noise ratio (SNR) conditions, scenarios that are common in real-world hyperspectral imagery, thereby limiting their robustness and practical applicability. To address this, we propose \textbf{WS-Net}, a novel deep learning framework specifically designed for hyperspectral unmixing in weak signal scenario. Our contributions are as follows:

\begin{itemize}
\item To address weak signals at feature extraction, we introduce a dual-branch encoder combining wavelet-based frequency decomposition (Haar and Symlet-3) with convolutional spatial processing, enabling enhanced representation of low-magnitude spectral features while suppressing high-frequency noise.

\item To efficiently capture long-range spectral–spatial dependencies without suppressing weak spectra, we design a weak-signal aware State Space Model (SSM) and Transformer fusion module. It integrates a Mamba-based state-space branch that linearly propagates spectral context with a weak-signal attention branch that reweights low-similarity tokens through inverse attention, while a learnable gate adaptively fuses both branches for robust weak-endmember recovery.

\item To promote accurate abundance estimation for low-reflectance materials, we propose a weak signal-aware decoder that incorporates softmax-constrained abundance mapping and a KL divergence regularisation term, encouraging spectral disentanglement between weak and dominant endmembers.


\end{itemize}

Across a synthetic benchmark and two real datasets, WS-Net consistently excels under realistic remote-sensing conditions, particularly when low-reflectance or small-fraction endmembers dominate. On the Synthetic dataset, it achieves the lowest errors, reducing RMSE relative to FCLSU and MiSiCNet~\cite{rasti2022misicnet} by 36\% and 55\%, and lowering SAD relative to FCLSU and DeepTrans~\cite{9848995} by 73\% and 63\%. On Samson~\cite{zhu2017hyperspectral}, it attains the best mean SAD, indicating more accurate abundance directions. On Apex~\cite{delalieux2014unmixing,zhu2017hyperspectral}, it reaches the best overall performance, with the largest gains on weak-signal classes such as Road and Water, and the best SAD across all four endmembers.



\section{Related Work}

\subsection{Traditional Spectral Unmixing Techniques}
Most classical hyperspectral unmixing (HU) methods adopt the Linear Mixing Model (LMM), which assumes each observed pixel spectrum is a convex combination of multiple endmember spectra weighted by their abundance fractions~\cite{keshava2002spectral,chang2010review}. Given known or pre-estimated endmembers, abundance estimation is commonly posed as a constrained optimization that enforces the Abundance Non-negativity Constraint (ANC) and the Abundance Sum-to-One Constraint (ASC) to ensure physical interpretability~\cite{chang2010review}. Representative algorithms include Fully Constrained Least Squares Unmixing (FCLSU)~\cite{heinz1999fully} and Non-negative Matrix Factorisation (NMF) variants~\cite{li2012collaborative}.
When endmembers are unknown, blind or unsupervised unmixing jointly estimates endmembers and abundances by injecting geometric or statistical priors~\cite{borsoi2021spectral}. Methods such as Vertex Component Analysis (VCA)~\cite{nascimento2005vertex} exploit simplex geometry under the pure-pixel assumption, while minimum-volume–constrained NMF~\cite{li2012collaborative} promotes a compact endmember simplex to alleviate the absence of pure pixels. To improve robustness, recent work has explored transformed domains (e.g., wavelets) to enhance denoising and separability before or during unmixing~\cite{9555396}. Although computationally efficient and interpretable, these approaches inherit LMM linearity assumptions and often degrade in noise-prone or non-Gaussian environments, where standard least-squares losses become suboptimal and estimation bias increases~\cite{chan2024sparse}.

A particularly challenging scenario arises when certain materials exhibit inherently low reflectance across the spectral range. These low-reflectance endmembers, such as water, shaded regions, or trace contaminants, produce weak signals that are easily dominated by stronger components or corrupted by noise. Consequently, traditional HU methods struggle to accurately detect and quantify such endmembers, resulting in abundance underestimation or complete omission. This limitation significantly impacts the reliability and applicability of HU models in real-world scenarios where weak signals play a critical role.

\subsection{Deep Learning for Hyperspectral Unmixing}
With the advancement of data-driven methodologies, deep learning has emerged as a powerful tool for HU, enabling the modelling of complex spectral–spatial correlations and the learning of hierarchical representations~\cite{lecun2015deep,bhatt2020deep}. Early efforts such as mDAU~\cite{guo2015hyperspectral} utilised autoencoder-based architectures with denoising and sparsity constraints to extract robust latent features. However, these methods mainly focused on spectral reconstruction and lacked explicit spatial modelling. To incorporate spatial priors, subsequent approaches introduced convolutional neural networks. For example, EndNet~\cite{ozkan2018endnet} employed ReLU activations and spectral angle distance (SAD) minimisation, while CNNAEU~\cite{palsson2020convolutional} used convolutional layers to encode spatial context. CNNAEU2 further added a softmax activation to enforce the abundance sum-to-one constraint (ASC), improving the physical interpretability of abundance maps.

Despite these advances, most deep HU models either focus on spectral information or capture spatial features via shallow local filters, making them insufficient to model long-range dependencies. MiSiCNet~\cite{rasti2022misicnet} introduced a hybrid structure that combines CNN-based spatial features with a minimum simplex volume regulariser, enhancing robustness in noisy and pure-pixel-absent scenarios. Nevertheless, the architecture remains inadequate for detecting low-reflectance endmembers, whose weak and sparse spectral signatures are often overwhelmed by dominant signals or noise.

To address this issue, Transformer-based architectures have been recently explored. DeepTrans-HSU~\cite{9848995} integrates CNN encoders with Multi-Head Self-Attention (MHSA) to model long-range dependencies across pixels. While MHSA significantly improves representational capacity, it incurs high computational costs and tends to dilute weak signal features due to global averaging operations. However, the MHSA are primarily designed to emphasise dominant responses, and they remain suboptimal in scenarios involving low signal-to-noise ratio or spatially sparse endmembers.

Recent state-space sequence models such as S4~\cite{dao2024mamba}, HyMix-SSM~\cite{glorioso2024zamba}, and MambaVision~\cite{Hatamizadeh_2025_CVPR_mambavision} have demonstrated efficient global context modelling in computer vision and have great potential for HU. However, these architectures are largely agnostic to hyperspectral physics; they neither enforce abundance conservation nor address spectral imbalance between dominant and weak components. Our proposed WS-Net extends this paradigm through a dual adaptation; one branch is dedicated to high-energy reconstruction, and another to long-term weak-signal retention, allowing it to operate under the abundance and non-negativity constraints intrinsic to hyperspectral unmixing.


\subsection{Toward Weak Signal Hyperspectral Unmixing}
Most methods assumes key endmembers are locally spectrally dominant and spatially continuous in local regions.Most methods assumes key endmembers are locally spectrally dominant and spatially continuous in local regions.  (e.g., dark minerals, shaded water bodies, trace materials)~\cite{shen2022toward}. These weak signals are easily overwhelmed during feature extraction, especially in attention or pooling-based architectures. Recent work, such as HyperWeak~\cite{shen2022toward}, addressed this by injecting prior knowledge into matrix factorisation frameworks, but suffered from optimisation complexity and poor scalability to deep architectures.

Furthermore, weak endmembers face dual challenges: (1) low spectral energy, which makes them vulnerable to suppression or aliasing, and (2) sparse, scattered spatial distribution, complicating their detection and reconstruction. As a result, most current models exhibit what we refer to as \textit{weak signal collapse}, where low-reflectance materials are either underrepresented or entirely omitted in the final abundance maps.

\noindent\textbf{Our Positioning:}  
This work addresses the above gap by introducing a deep unmixing architecture, WS-Net, that explicitly models weak signals across both spectral and spatial domains. As detailed in Section~\ref{sec:intro}, the proposed WS-Net framework is purpose-built for weak signal hyperspectral unmixing. It integrates multi-resolution wavelet encoding, dual state-space modelling across spatial and spectral domains, and a sparsity-aware decoder with KL divergence regularisation to effectively enhance and recover low-reflectance endmembers under challenging conditions.

\begin{figure*}[htbp]
    \centering
    \includegraphics[width=1\textwidth]{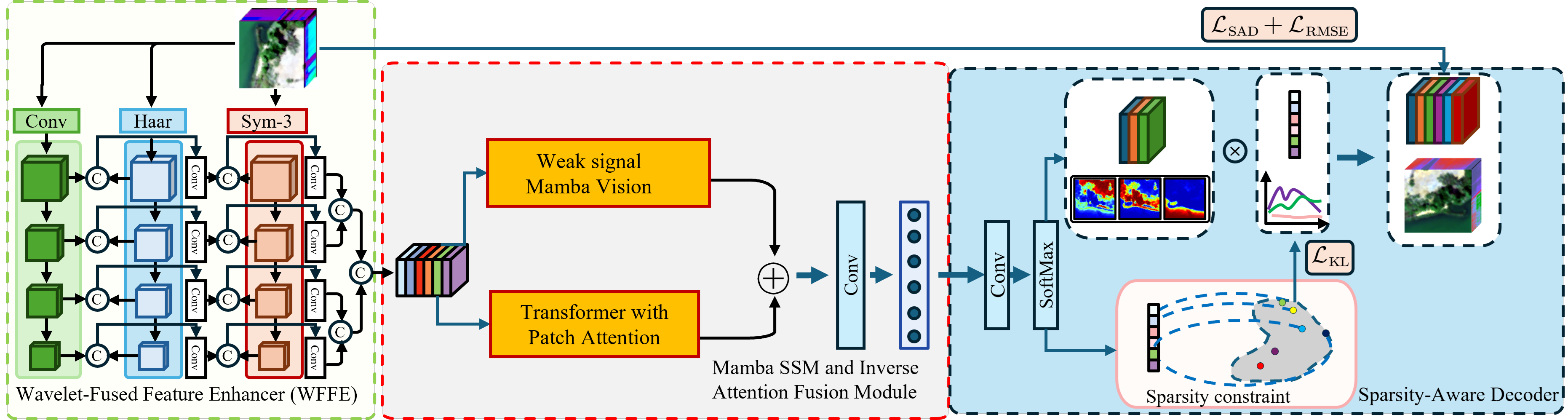}
\caption{Overview of the proposed WS-Net framework. 
(1) \textbf{WFFE}: A multi-resolution wavelet-fused module that captures spatial–spectral cues across scales and enhances low-reflectance signals via Haar and Symlet-3 convolutions. 
(2) \textbf{Mamba SSM and Inverse Attention Fusion}: A dual-branch backbone integrating Mamba state-space modelling for efficient long-range spectral propagation with inverse attention for selective weak-signal enhancement. 
(3) \textbf{Sparsity-Aware Decoder}: Reconstructs hyperspectral data from softmax-constrained abundance maps while enforcing sparsity and distributional separation through KL-divergence regularisation.}
    \label{fig:overall}
\end{figure*}
\section{Theoretical Formulation of Weak Signal Collapse in Hyperspectral Unmixing}\label{sec:problem_formulation}


Weak-signal unmixing can be regarded as an ill-posed nonlinear inverse problem that is often approximated by a locally linear model for analytical tractability. 
To provide a baseline formulation, we first recall the standard linear mixing model (LMM). 
Let \(\mathbf{Y} \in \mathbb{R}^{L \times N}\) represent a hyperspectral image with \(L\) spectral bands and \(N\) total pixels. 
The LMM assumes that each pixel is a convex combination of a finite set of endmember spectra plus noise:
\begin{equation}
    \mathbf{Y} = \mathbf{E}\,\mathbf{A} + \mathbf{W}.
\label{function:HSI_Linear_model}
\end{equation}
Here, \(\mathbf{E} \in \mathbb{R}^{L \times P}\), \(\mathbf{A} \in \mathbb{R}^{P \times N}\), and \(\mathbf{W} \in \mathbb{R}^{L \times N}\) denote the endmember, abundance, and additive noise matrices, respectively, with \(P\) the number of endmembers. 
The LMM is constrained by three well-known physical conditions:

\begin{itemize}
    \item \textbf{Abundance Nonnegativity Constraint (ANC):} \(a_{ij} \ge 0\) for all abundance coefficients \(a_{ij}\) in \(\mathbf{A}\).
    \item \textbf{Abundance Sum-to-One Constraint (ASC):} For each pixel \(n=1,\ldots,N\), \(a_{1,n} + a_{2,n} + \cdots + a_{P,n} = 1\).
    \item \textbf{Full Column Rank:} The rank condition \(\min(L,N) > P\) holds, and \(\mathbf{E}\) is of full column rank.
\end{itemize}

While Eq.~\eqref{function:HSI_Linear_model} provides a convenient first-order approximation, 
real hyperspectral mixtures often deviate from strict linearity due to scattering, illumination geometry, and inter-material coupling effects that become particularly pronounced for weak-reflectance endmembers. 
Thus, the observed spectrum can be more generally expressed as
\begin{equation}
    \mathbf{Y} = \mathcal{F}(\mathbf{E}, \mathbf{A}) + \mathbf{W}
    = \mathbf{E}\,\mathbf{A} + \boldsymbol{\Delta}(\mathbf{E}, \mathbf{A}) + \mathbf{W},
\label{eq:nonlinear_model}
\end{equation}
where \(\mathcal{F}(\cdot)\) is a nonlinear observation operator and \(\boldsymbol{\Delta}(\mathbf{E}, \mathbf{A})\) represents the residual nonlinear interactions unaccounted for by the LMM.
When signal magnitudes are small, \(\|\boldsymbol{\Delta}(\mathbf{E}, \mathbf{A})\|\) is non-negligible, making the inversion from \(\mathbf{Y}\) to \(\mathbf{A}\) highly ill-posed.

In practical unmixing pipelines, the nonlinear term is often ignored and the LMM is used as a working approximation. 
However, for low-reflectance materials, this approximation breaks down, leading to what we term \emph{weak-signal attenuation}: the systematic underestimation of endmembers whose contributions are faint or spatially sparse.

To explicitly account for such endmembers, we partition the endmember matrix \(\mathbf{E}\) into two subsets:
\[
\mathbf{E} = \bigl[\, \mathbf{E}_s \;\; \mathbf{E}_w \bigr],
\]
where, \(\mathbf{E}_s \in \mathbb{R}^{L \times p_s}\) denotes standard, moderate-to-high reflectance endmembers, and \(\mathbf{E}_w \in \mathbb{R}^{L \times p_w}\) contains endmembers with low reflectance or nonlinear spectral distortion. The abundance matrix is partitioned accordingly:
\[
\mathbf{A} = 
\begin{bmatrix}
\mathbf{A}_s \\[4pt]
\mathbf{A}_w
\end{bmatrix},
\]
where \(\mathbf{A}_s \in \mathbb{R}^{p_s \times N}\) and \(\mathbf{A}_w \in \mathbb{R}^{p_w \times N}\) are the abundance fractions of the respective subsets. 
Substituting these into the general model yields:
\begin{align}
    \mathbf{Y} 
    &= 
    \bigl[\, \mathbf{E}_s \;\; \mathbf{E}_w \bigr]
    \begin{bmatrix}
        \mathbf{A}_s \\
        \mathbf{A}_w
    \end{bmatrix}
    + \boldsymbol{\Delta}(\mathbf{E}, \mathbf{A}) + \mathbf{W} \notag\\
    &=
    \mathbf{E}_s \mathbf{A}_s + \mathbf{E}_w \mathbf{A}_w
    + \boldsymbol{\Delta}(\mathbf{E}, \mathbf{A}) + \mathbf{W}.
\label{eq:extended_LMM}
\end{align}


Equation~\eqref{eq:extended_LMM} emphasises that the weak-signal component \(\mathbf{E}_w \mathbf{A}_w\) is not only small in energy but also prone to nonlinear distortions captured by \(\boldsymbol{\Delta}(\mathbf{E}, \mathbf{A})\). 
This term renders the estimation of \(\mathbf{A}_w\) unstable and motivates the use of learning-based regularisation to approximate \(\boldsymbol{\Delta}(\cdot)\) while preserving the linear abundance constraints. 
The proposed WS-Net architecture in Section~\ref{sec:framework} can thus be viewed as a data-driven nonlinear operator that compensates for \(\boldsymbol{\Delta}(\cdot)\), restoring weak-endmember information under low-SNR conditions.

\section{WS-Net Framework for Weak Signal Hyperspectral Unmixing}
\label{sec:framework}

We propose WS-Net, a deep neural architecture for end-to-end hyperspectral unmixing under weak-signal conditions. The model integrates three key components: (i) a multi-resolution wavelet-based encoder that extracts discriminative spatial–spectral features while suppressing high-frequency noise, (ii) a dual-branch backbone combining state-space modelling (Mamba) and transformer-based global context encoding, and (iii) a sparsity-aware decoder that reconstructs the hyperspectral data and estimates abundance maps under physical constraints.
An overview of the complete pipeline is illustrated in Fig.~\ref{fig:overall}, which outlines the flow of information through each component.

\subsection{Wavelet-Fused Spectral-Spatial Feature Extractor (WFFE)}
\label{subsection:wfsse}

The WFFE module is designed to enhance hyperspectral feature extraction by capturing both multi-resolution spatial textures and spectral continuity, which are critical for identifying low-reflectance materials. Traditional convolutional encoders often fail to preserve subtle variations in weak signals, particularly in the presence of noise. To address this, we integrate wavelet decomposition into the encoder pathway.

 We employ two types of discrete wavelet transforms (DWT): Haar and Symlet-3~\cite{anand2021wavelet,guo2022hyperspectral}. These are applied to the input hyperspectral cube to extract spatial features at different frequency bands. Haar captures coarse-scale structures, while Symlet-3 preserves smoother transitions and finer spatial details. The resulting coefficients are then fused to create a composite representation that preserves weak edge structures and localised variations, key indicators of low-reflectance endmembers.


Specifically, the WFFE module first applies a two-dimensional discrete wavelet transform (2D-DWT) along the spatial dimension of the hyperspectral image, generating a set of low-frequency (approximation) and high-frequency (detail) sub-bands. The low-frequency sub-bands preserve the dominant spatial structures and background spectral information, whereas the high-frequency sub-bands capture textures, edges, noise, and subtle spectral variations that are inherent to weak signals. This decomposition enables differentiated processing of distinct frequency components, thereby retaining the global information of strong signals while selectively enhancing weak spectral responses embedded in the high-frequency details.

As illustrated in Fig.~\ref{fig:overall},  the WFFE consists of four stacked stages. In each stage, the input feature map is decomposed using both the Haar~\cite{guo2022hyperspectral} and Symlet-3 wavelet transforms~\cite{anand2021wavelet,guo2022hyperspectral} as follows:
\begin{equation}
    F^{i} = \text{Conv}(F^{i-1}_\text{Haar} \text{\textcircled{c}} F^{i-1}_\text{conv}) \text{\textcircled{c}} F^{i-1}_\text{Symlet-3}, \hspace{0.5cm} i=1,2,3,4
\end{equation}
where $i$ is the stage number. \textcircled{c} denotes concatenation processing. The Haar basis captures sharp edges and discontinuities, while Symlet-3, with higher vanishing moments and smoother support, models continuous, low-amplitude variations. We further apply a feature fusion by a convolution-based resizing to align multi-scale features:
\begin{equation}
\begin{split}
    \widetilde{F}^{i} &= \text{Conv}(F^{i}), \quad i = 1, 2, 3, 4 \\
    F &= \widetilde{F}^{1} \text{\textcircled{c}} \widetilde{F}^{2} \text{\textcircled{c}} \widetilde{F}^{3} \text{\textcircled{c}} \widetilde{F}^{4}
\end{split}
\end{equation}
where a series of convolutional layers transform feature maps from multiple resolutions into a shared feature space, providing adaptive transformation capabilities to ensure semantic consistency across scales. The resulting feature cube $F$ is then fed into the encoder for subsequent high-level representation learning. 

By integrating Haar and Symlet-based multi-frequency features across multiple stages into a spatially aligned representation, WFFE significantly improves the model’s ability to capture both sharp boundaries and weak reflectance patterns in real-world HSI data.

Following wavelet fusion, a series of 2D convolutional layers are applied to jointly learn spatial and spectral representations. This fused tensor forms a robust input to the dual-backbone architecture described in Section IV-B, enhancing the network’s ability to disentangle overlapping spectral features while suppressing high-frequency noise.



\begin{figure*}[thbp]
    \centering
    \includegraphics[width=0.85\textwidth]{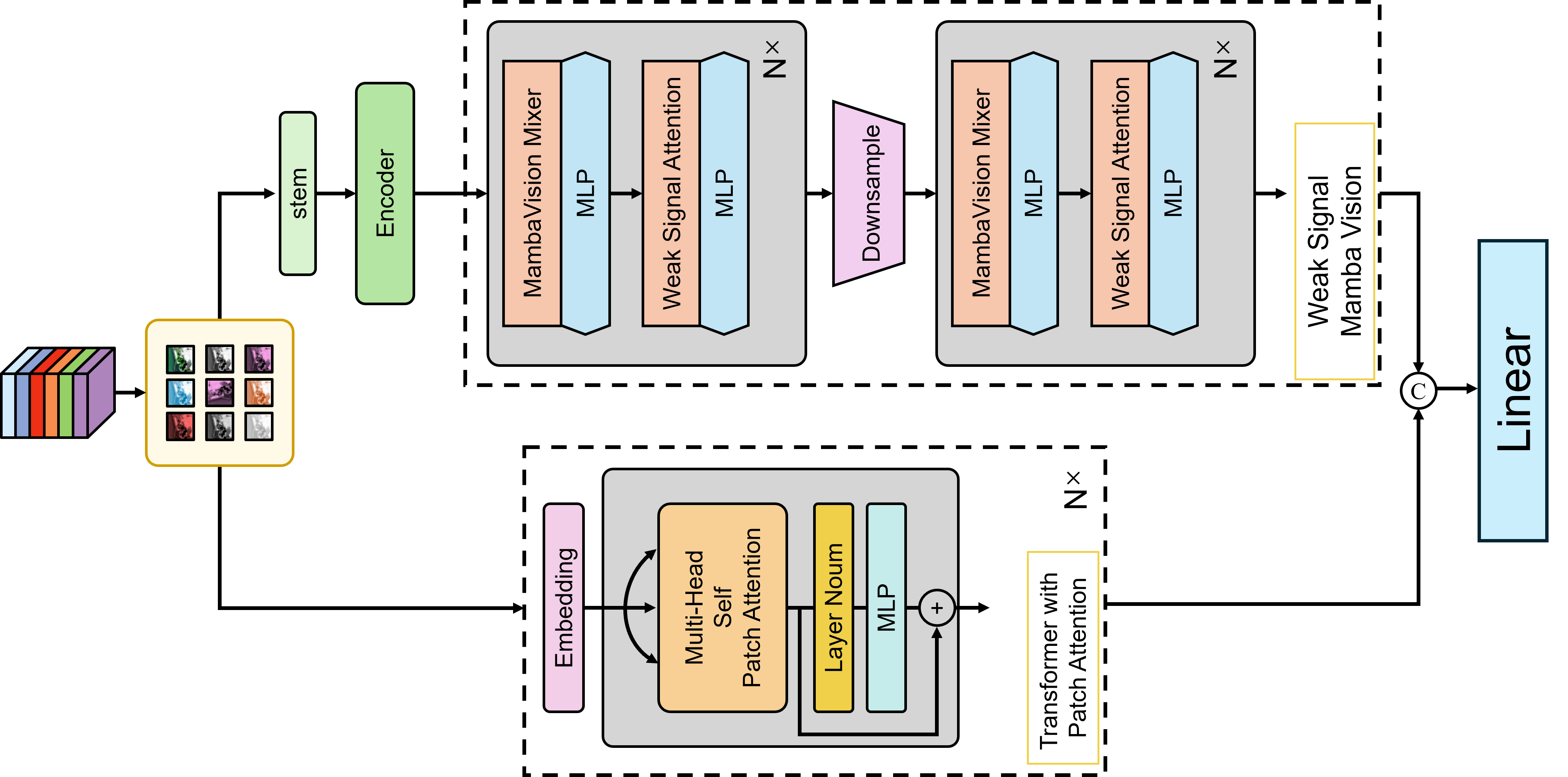}
\caption{Architecture of the proposed Mamba SSM and Inverse Attention Fusion module. 
The framework integrates a Mamba-based state-space branch for long-range spectral modelling with a weak-signal Weak Signal Attention branch that selectively amplifies low-energy features suppressed by noise. 
The Transformer pathway introduces a class token with patch-level attention to provide global contextual guidance for hyperspectral unmixing.}
    \label{fig:weak mamaba}
\end{figure*}


\subsection{Mamba SSM and Weak Signal Attention Fusion for Weak-Signal Modelling}
\label{section:Weak-Signal-Aware Mamba}

We adopt a hybrid design that couples a Mamba state-space branch with a Transformer-style Weak Signal Attention branch rather than duplicating either component. 
This architecture leverages their complementary modelling biases: the Mamba branch captures smooth, long-range spectral evolution through linear recurrent dynamics with low computational cost, while the Weak Signal Attention branch emphasises sparse, high-contrast cues typically associated with weak or low-abundance endmembers. 
Using the Mamba modules would efficiently propagate global context but fail to selectively amplify weak responses, whereas the Transformers would increase selectivity at the expense of complexity and numerical stability. 
The proposed SSM–Transformer fusion therefore combines continuous spectral dynamics and discrete contextual reweighting within a single weak-signal–aware framework. 
Unlike conventional architectures that encode all features into a unified representation, this explicit separation enables the model to address weak-signal degradation from both local and global perspectives, leading to more stable and accurate unmixing under low-SNR conditions. 
An overview of the dual-branch framework is shown in Fig.~\ref{fig:weak mamaba}.

\subsubsection{Mamba State-Space Branch}
The raw hyperspectral image is first processed by the \textit{Wavelet-Fused Spectral–Spatial Feature Extractor} to obtain a sequence representation:
\[
\mathbf{Y}' \in \mathbb{R}^{L' \times N'},
\]
where $L'$ denotes the compressed spectral dimension and $N'$ the total number of spatial tokens. 
Both branches operate on identical token embeddings, enabling the network to learn complementary representations of the same spectral–spatial context.

In the Mamba branch, the sequence $\mathbf{Y}'$ is processed using a structured state-space model (SSM) following the Mamba architecture. 
The SSM replaces self-attention with a learned state-update mechanism that captures long-range dependencies through recurrent operations while maintaining linear time and space complexity. 
Each Mamba block applies gated convolutions and dynamic weighting to isolate weak, noise-obscured signals that may not exhibit strong token similarity. 
The branch thus preserves faint signatures across neighbouring pixels and spectral bands, providing a noise-robust local reconstruction of low-SNR components.

\subsubsection{Weak Signal Attention Branch}
To model global dependencies and emphasise weak but informative spectral cues, a Transformer-based branch is employed. 
Each token is treated as a patch embedding augmented with a learnable positional encoding to capture spatial structure. 
A linear projection maps $\mathbf{Y}'$ to the query, key, and value matrices:
\[
\mathbf{Q} = \mathbf{Y}'\mathbf{W}_Q, \quad
\mathbf{K} = \mathbf{Y}'\mathbf{W}_K, \quad
\mathbf{V} = \mathbf{Y}'\mathbf{W}_V,
\]
Following the standard scaled dot-product attention formulation~\cite{vaswani2017attention}, the similarity matrix is calculated as:
\[
\mathbf{S} = \frac{\mathbf{Q}\mathbf{K}^{\top}}{\sqrt{d_k}},
\]
where $d_k$ denotes the head dimension.

To address the under-emphasis of weak yet informative responses, we introduce a \textit{Weak-Signal Attention (WSA)} mechanism with two complementary branches. 
The first follows the standard attention formulation:
\begin{equation}
\mathbf{A}_{\mathrm{sa}} = \mathrm{Softmax}\!\left(\frac{\mathbf{S}}{\tau_{\mathrm{sa}}}\right), 
\quad
\mathbf{Z}_{\mathrm{sa}} = \mathbf{A}_{\mathrm{sa}}\mathbf{V},
\end{equation}
which aligns highly similar tokens and stabilises dominant spectral patterns.
The second, termed \textit{Normalised Inverse Attention (NIA)}, redistributes probability mass toward low-similarity pairs, thereby amplifying weak spectral responses:
\begin{equation}
\mathbf{A}_{\mathrm{inv}} = \mathrm{Softmax}\!\left(-\frac{\mathbf{S}}{\tau_{\mathrm{inv}}}\right),
\quad
\mathbf{Z}_{\mathrm{inv}} = \mathbf{A}_{\mathrm{inv}}\mathbf{V},
\end{equation}
where $\tau_{\mathrm{sa}}$ and $\tau_{\mathrm{inv}}$ control attention sharpness.

A learnable gate $\alpha = \sigma(\mathrm{gate})$ adaptively fuses both attention maps. Here, $\sigma(\cdot)$ denotes the sigmoid activation, ensuring $\alpha \in [0,1]$. 
The gating parameter $\alpha$ is produced by a lightweight feedforward network  conditioned on intermediate attention activations, enabling adaptive weighting between the two attention maps:
\begin{equation}
\mathbf{A}_{\mathrm{fuse}} = \alpha\,\mathbf{A}_{\mathrm{sa}} + (1 - \alpha)\,\mathbf{A}_{\mathrm{inv}}, \qquad
\mathbf{Z}_{\mathrm{Attention}} = \mathbf{A}_{\mathrm{fuse}}\mathbf{V}.
\end{equation}
The gating network is implemented as a lightweight feedforward layer conditioned on intermediate activations, enabling per-sample adaptation. 
When the signal-to-noise ratio (SNR) is low, $\alpha$ increases the contribution of the Weak Signal Attention branch to enhance weak responses; under stable conditions, it favours standard attention to preserve strong alignments. 

\subsubsection{Fusion}
Finally, $\mathbf{Z}_{\mathrm{Attention}}$ is concatenated with the Mamba output along the channel dimension and passed through a token-wise multilayer perceptron (MLP), functionally equivalent to the Transformer feed-forward network (FFN), to perform nonlinear reweighting and feature refinement. 
This design preserves the probabilistic interpretation of attention and maintains compatibility with multi-head configurations, allowing seamless integration with Transformer blocks. 
To balance the complementary contributions of the Mamba and attention branches, a learnable global gating mechanism adaptively fuses their outputs:
\begin{equation}
\mathbf{Z}_{\text{final}} = \beta\,\mathbf{Z}_{\text{Mamba}} + (1 - \beta)\,\mathbf{Z}_{\text{Attention}},
\end{equation}
where $\beta = \sigma(\mathrm{gate}_{\text{global}})$ is a learnable gate that adaptively
weights local state-space dynamics against global contextual cues.
This global fusion allows the network to rely more on the Mamba branch under high-SNR conditions
and shift toward the attention branch when weak or noisy signals dominate, 
achieving a self-balancing mechanism that unifies continuous and discrete spectral modelling.
Both $\mathbf{Z}_{\text{Mamba}}$ and $\mathbf{Z}_{\text{Attention}}$ are projected to the same feature dimension to ensure a valid element-wise fusion.
The fused representation $\mathbf{Z}_{\text{final}}$ is then passed to the \textit{Sparsity-Aware Decoder} for abundance estimation and hyperspectral reconstruction.

\subsection{Sparsity-Aware Decoder: Losses and Optimisation Functions}
\label{subsec:kl_sparse_unmixing}


To strengthen the sparsity and robustness of weak-signal reconstruction, the Sparsity-Aware Decoder maps the fused latent features to abundance maps that satisfy the physical abundance constraints. A softmax activation along the endmember dimension enforces both non-negativity and the \textit{abundance-sum-to-one constraint (ASC)}.
Let $\mathbf{Z} \in \mathbb{R}^{R \times H \times W}$ denote the latent feature tensor, where $R$ is the number of endmembers.
The abundance maps are computed as:
\begin{equation}
\mathbf{A}_{r,h,w} = \frac{\exp(Z_{r,h,w})}{\sum_{r'=1}^{R}\exp(Z_{r',h,w})},
\end{equation}
ensuring $\mathbf{A}_{r,h,w} \ge 0$ and $\sum_r \mathbf{A}_{r,h,w} = 1$.
The final reconstruction is obtained as:
\begin{equation}
\hat{\mathbf{Y}} = \mathbf{E}\mathbf{A},
\end{equation}
where $\mathbf{E} \in \mathbb{R}^{L \times R}$ represents the endmember matrix.
In practice, $\mathbf{E}$ is initialised using endmembers extracted by the VCA algorithm~\cite{nascimento2005vertex} and fine-tuned jointly during training, allowing the decoder to refine spectral bases for both weak and dominant materials.

The Sparsity-Aware Decoder is trained using a composite loss that integrates Spectral Angle Distance (SAD), Root Mean Square Error (RMSE), and Kullback–Leibler (KL) divergence. This combination enforces complementary constraints, energy fidelity, directional consistency, and distributional separability, optimising the decoder for reliable abundance estimation under low-reflectance and high-variability conditions. The joint objective ensures that weak and dominant endmembers are reconstructed with distinct spectral profiles while maintaining accurate pixel-level reconstruction, which is critical for unmixing datasets characterised by unstable or noise-prone spectral patterns.

To promote spectral discriminability between strong and weak endmembers, a KL-divergence-based regularisation strategy is introduced~\cite{Mantripragada2024KL,palsson2022blind}. This approach imposes a distributional constraint on normalised spectral features, explicitly encouraging weak endmembers to occupy distinct and separable spectral subspaces. Unlike energy-based losses, KL divergence evaluates the relative distribution of the entire spectral curve rather than absolute reflectance magnitudes. This property enables the model to focus on capturing the intrinsic spectral shape and patterns of weak signals, thereby mitigating the influence of noise and wavelength instabilities and ensuring robust reconstruction.

Let $\mathbf{Y} \in \mathbb{R}^{H \times W \times L}$ denote the ground-truth hyperspectral image and $\hat{\mathbf{Y}} \in \mathbb{R}^{H \times W \times L}$ the reconstructed counterpart. For each pixel $(i, j)$, $\mathbf{e}_{ij} \in \mathbb{R}^{L}$ and $\hat{\mathbf{e}}_{ij} \in \mathbb{R}^{L}$ represent the ground-truth and reconstructed spectral vectors, respectively. Each vector is normalised into a probability distribution as follows:
\[
P(\mathbf{e}_{ij}) = \frac{\mathbf{e}_{ij}}{\sum_{l=1}^{L} e_{ij,l}}, \quad 
\hat{P}(\hat{\mathbf{e}}_{ij}) = \frac{\hat{\mathbf{e}}_{ij}}{\sum_{l=1}^{L} \hat{e}_{ij,l}} .
\]
The KL divergence between the reconstructed and true spectra is defined as:
\begin{equation}
L_{\text{KL}} = \frac{1}{H \cdot W} \sum_{i=1}^{H} \sum_{j=1}^{W} D_{\text{KL}} \left( P(\mathbf{e}_{ij}) \,\|\, \hat{P}(\hat{\mathbf{e}}_{ij}) \right),
\end{equation}
where $D_{\text{KL}}(P \| Q) = \sum_{l=1}^{L} P_l \log \tfrac{P_l}{Q_l}$.

The KL divergence encourages the decoder to preserve the \textit{spectral shape} of each reconstructed pixel rather than its absolute magnitude.
By normalising $\mathbf{e}_{ij}$ and $\hat{\mathbf{e}}_{ij}$ into probability distributions across spectral bands,
$L_{\mathrm{KL}}$ measures divergence in the \textit{relative energy distribution} of spectra.
This regulariser is particularly beneficial for weak endmembers, whose absolute reflectance values are small but whose spectral shapes remain distinctive.
To prevent numerical instability when any spectral element approaches zero,
a small constant $\varepsilon = 10^{-8}$ is added during normalisation and logarithmic computation. This stabilises optimisation and ensures that the KL term prioritises spectral pattern fidelity over raw intensity matching.

In addition to KL divergence, we incorporate two reconstruction-driven losses. The first is the  Root Mean Square Error (RMSE), which measures pixel-wise deviations between the original and reconstructed hyperspectral images:
\begin{equation}
L_{\text{RMSE}}(\mathbf{Y}, \hat{\mathbf{Y}}) = 
\sqrt{\frac{1}{H \cdot W \cdot L} 
\sum_{i=1}^{H} \sum_{j=1}^{W} \sum_{l=1}^{L} \left( \hat{Y}_{ijl} - Y_{ijl} \right)^2 } .
\end{equation}

The second is the Spectral Angle Distance (SAD), which enforces directional similarity between ground-truth and reconstructed spectra, thereby preserving spectral shape consistency:
\begin{equation}
L_{\text{SAD}}(\mathbf{Y}, \hat{\mathbf{Y}}) = 
\frac{1}{H \cdot W} \sum_{i=1}^{H} \sum_{j=1}^{W} 
\arccos \left( \frac{ \langle \mathbf{e}_{ij}, \hat{\mathbf{e}}_{ij} \rangle }{ \| \mathbf{e}_{ij} \|_2 \cdot \| \hat{\mathbf{e}}_{ij} \|_2 } \right).
\end{equation}

The final objective function integrates all three components to ensure not only accurate reconstruction of hyperspectral scenes but also the learning of sparse and distinctive spectral constituents, which is critical for robust identification of low-reflectance endmembers:
\begin{equation}
L = \alpha_l L_{\text{RMSE}} + \beta_l L_{\text{SAD}} + \gamma_l L_{\text{KL}} ,
\end{equation}
where $\alpha_l,\beta_l,\gamma_l>0$ balance energy fidelity, angular consistency, and endmember-distribution regularisation. This formulation is particularly effective in weak-signal regimes with spectral variability, yielding robust endmember and abundance estimates.

\section{ Experimental Setup}

\subsection{Datasets}
\label{Datasets}
\begin{figure*}[htbp]
    \centering
    \includegraphics[width=\textwidth]{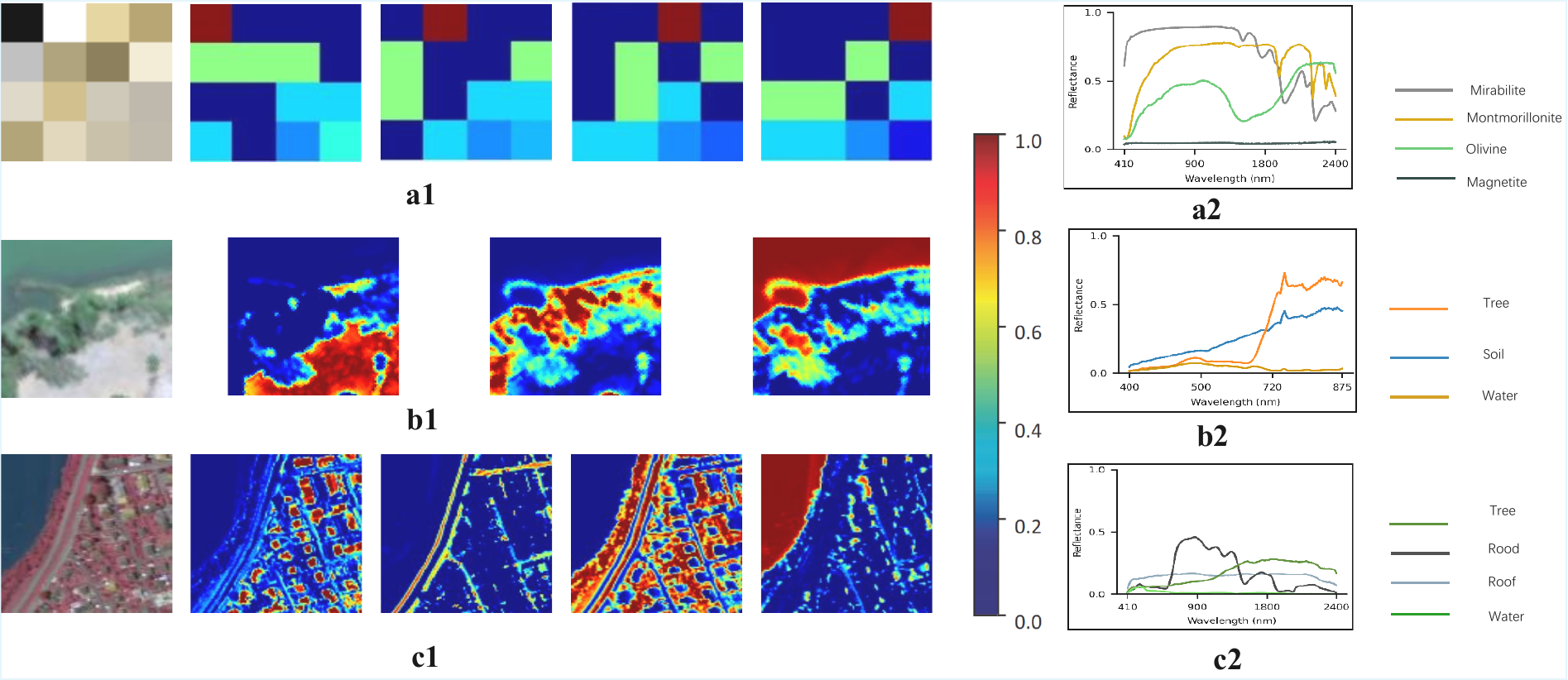}
   \caption{Dataset visualizations: (\textbf{a})~Simulated dataset, (\textbf{b})~Samson dataset, (\textbf{c})~Apex dataset. 
For each dataset, we show (left) a pseudo-RGB image (R: 650\,nm, G: 532\,nm, B: 450\,nm), 
(middle) representative endmember abundance maps (normalized; brighter indicates higher abundance), 
and (right) endmember spectral signatures (reflectance versus wavelength).}

    \label{fig:data_vis}
\end{figure*}

To comprehensively evaluate our method, we conduct experiments on one synthetic and two widely used real-world hyperspectral datasets: \textbf{Simulated}, \textbf{Samson}, and \textbf{Apex}. In this study, a \textbf{weak signal} endmember refers to a material whose spectral reflectance is substantially lower than that of other materials across most wavelengths within the same scene. Such low-reflectance endmembers typically yield fewer photon returns and exhibit weaker separability in the spectral domain, thus complicating the unmixing process. 
Pseudo-RGB composites, abundance maps, and endmember spectral curves for each dataset are shown in Fig.~\ref{fig:data_vis}.

\subsubsection{Simulated Dataset with Low-Reflectance Endmember}
We construct a synthetic dataset comprising \(80 \times 80\) pixels by linearly mixing four spectral signatures selected from the USGS spectral library~\cite{kokaly2017usgs}. The chosen materials include magnetite, mirabilite, montmorillonite, and olivine, corresponding to the spectral files \textit{magnetite},\textit{ mirabilite}, \textit{montmorillonite}, and \textit{olivine}, respectively. This image contains 16 squares of 20 × 20 pixels with different ternary mixtures.

Each hyperspectral pixel contains 459 reflectance values uniformly sampled in the wavelength range from 400~nm to 2500~nm. Spatially, the image is composed of 16 non-overlapping square regions of \(20 \times 20\) pixels, each representing a distinct abundance composition. Abundance vectors are randomly generated to satisfy the abundance nonnegativity and sum-to-one constraints, following the linear mixing model.

In this dataset, \textbf{magnetite} is designated as the \textit{weak-signal endmember} due to its extremely low reflectance (typically $<$0.1) across most of the visible to shortwave infrared spectrum, as specified by the USGS database. 

\subsubsection{Samson Dataset}
The Samson dataset consists of \(95 \times 95\) pixels and 156 spectral bands in the wavelength range from 401~nm to 889~nm~\cite{zhu2014spectral,zhu2017hyperspectral}. It contains three primary materials: Soil, Tree, and Water. The ground truth endmember spectra are manually selected from representative regions in the image, following standard practice in hyperspectral unmixing studies. This dataset serves as a benchmark for evaluating unmixing performance in scenes with moderate reflectance contrast.

In this dataset, \textbf{water} serves as the \textit{weak-signal component} because its reflectance throughout the visible to near-infrared range is markedly lower (typically $<$0.05) than that of vegetation or soil surfaces. The low reflectivity is attributed to strong optical absorption and minimal surface scattering. Consequently, the Samson dataset provides a moderately challenging scenario where weak water signals coexist with medium- and high-reflectance materials, facilitating evaluation of weak-signal discrimination in a controlled yet realistic environment.

\subsubsection{Apex Dataset}

The Apex dataset includes $110 \times 110$ pixels with 285 spectral bands, covering the wavelength range from 413\,nm to 2420\,nm~\cite{delalieux2014unmixing,zhu2017hyperspectral}. It contains four main endmembers: Water, Tree, Road, and Roof. These endmembers are extracted directly from prior-knowledge-based regions in the image. The dataset is more challenging due to its high spectral resolution and diverse material composition.

This dataset represents a highly challenging environment due to its extended spectral range, high spatial resolution, and diverse material composition. Both \textbf{roof} and \textbf{water} exhibit \textit{weak-signal characteristics}, albeit for different physical reasons. Roof materials (e.g., asphalt or aged tiles) have low spectral intensity in the near- and shortwave-infrared regions due to carbonaceous absorption and surface roughness, while water maintains uniformly low reflectance across the entire spectrum. The coexistence of multiple weak-signal materials within a complex scene makes the Apex dataset an ideal benchmark for evaluating the sensitivity and robustness of the proposed weak-signal-aware unmixing framework.

The ground-truth fractional abundances for the real datasets are computed using the Fully Constrained Least Squares Unmixing (FCLSU) algorithm, where the endmember signatures are manually selected from representative pure pixels in the scenes. These selected spectra serve as ground-truth endmembers to ensure reliable abundance estimation under the linear mixing model.

\subsection{Baseline Models}
 The simulated dataset is specifically designed to include a low-reflectance endmember (magnetite) to validate the model’s robustness under weak signal conditions. For each dataset, the number of endmembers \( R \) is assumed to be known and fixed across all methods. Ground-truth abundance maps \( \mathbf{A} \in \mathbb{R}^{R \times H \times W} \) and endmember matrices \( \mathbf{E} \in \mathbb{R}^{L \times R} \) are used as references for evaluation.

We compare the proposed model against six representative unmixing baselines, including classical, sparse, and deep learning-based methods:

\begin{itemize}
    \item \textbf{FCLSU + VCA}~\cite{heinz1999fully, nascimento2005vertex}: A geometrical unmixing approach based on fully constrained least squares, with endmembers extracted using the Vertex Component Analysis (VCA) algorithm.
    \item \textbf{Hyperweak}~\cite{shen2022toward}: A weak-signal-oriented unmixing framework built upon sparse nonnegative matrix factorisation (NMF). It includes an unsupervised reconstruction term and a supervised fidelity term to jointly estimate endmember and abundance matrices while preserving prior knowledge. A reweighted sparsity regularisation, Robust-OSP initialisation, and Nesterov’s optimal gradient method are employed for optimisation.
    \item\textbf{CNNAEU}~\cite{palsson2020convolutional}: A convolutional autoencoder-based method that reduces the spectral dimension and learns latent representations for abundance estimation.
    \item \textbf{DeepTrans}~\cite{9848995}: A transformer-based unmixing model employing spectral attention to capture long-range band dependencies.
    \item \textbf{EDAA}~\cite{zouaoui2023entropic}: A deep spectral attention model integrating global contextual information for improved endmember discrimination.
    \item \textbf{EndNet}~\cite{ozkan2018endnet}: A deep neural architecture incorporating spectral normalisation and sparse activations to enhance unmixing precision.
    \item \textbf{MiSiCNet}~\cite{rasti2022misicnet}: A spatial–spectral consistency-aware deep model using low-rank constraints for robust unmixing in real-world scenarios.
\end{itemize}

During the training of \emph{WS-Net}, endmember initialisation is performed using the Vertex Component Analysis (VCA) algorithm, and the parameters are subsequently optimised via gradient descent with the Adam optimiser. To ensure stable convergence and optimal performance across different datasets, dataset-specific learning rates (LR) and iteration numbers are adopted. For the synthetic S1 dataset, the LR is set to $4 \times 10^{-3}$ with 200 iterations; for the Samson dataset, the LR is set to $6 \times 10^{-3}$ with 200 iterations; and for the Apex dataset, the LR is set to $9 \times 10^{-3}$ with 300 iterations. All experiments are conducted on an NVIDIA GeForce RTX 3090 GPU with an Intel(R) Core(TM) i7-10700K CPU @ 3.80GHz.
All models are trained and tested using the same training-validation split. For simulated data, ground-truth endmembers are known, while for real datasets Hyperparameters for each method are tuned based on validation performance.

\subsection{Evaluation Metrices}

To quantitatively assess unmixing performance, we adopt two standard metrics: the root mean squared error (RMSE) for abundance estimation, and the spectral angle distance (SAD) for endmember accuracy.

\paragraph{Abundance RMSE.}
Let \( \mathbf{A}, \hat{\mathbf{A}} \in \mathbb{R}^{R \times H \times W} \) denote the ground-truth and estimated abundance maps, respectively. The RMSE is defined as:
\begin{equation}
\text{RMSE} = \sqrt{ \frac{1}{RHW} \sum_{r,i,j} \left( \hat{A}_{r,i,j} - A_{r,i,j} \right)^2 }.
\end{equation}

\paragraph{Spectral Angle Distance (SAD).}
To evaluate the spectral similarity between the estimated and ground-truth endmembers, the SAD is computed as:
\begin{equation}
\text{SAD}(\mathbf{E}, \hat{\mathbf{E}}) = \frac{1}{R} \sum_{i=1}^{R} \arccos \left( \frac{ \langle e_{(i)}, \hat{e}_{(i)} \rangle }{ \| e_{(i)} \|_2 \cdot \| \hat{e}_{(i)} \|_2 } \right),
\end{equation}
where \( e_{(i)} \) and \( \hat{e}_{(i)} \) are the \(i\)-th column vectors of the ground-truth endmember matrix \( \mathbf{E} \) and the estimated matrix \( \hat{\mathbf{E}} \), respectively, and \( \langle \cdot, \cdot \rangle \) denotes the inner product. This formulation directly reflects the \textbf{angular spectral distortion} and is widely adopted for evaluating endmember fidelity.

\section{Results and Analysis}

We now evaluate \emph{WS-Net} on both simulated and real HSI benchmarks with low-reflectance scenarios. This section details the evaluation protocol, datasets, metrics, and statistical tests, followed by comparisons with classical and recent unmixing methods (FCLS+VCA, CNNAEU~\cite{palsson2020convolutional}, EDAA~\cite{zouaoui2023entropic}, EndNet, MiSiCNet~\cite{rasti2022misicnet}, DeepTrans-HSU~\cite{9848995}). We report quantitative results, visualize representative abundance maps and spectra, and provide robustness analyses. 


\begin{figure*}[htbp]
    \centering
    \includegraphics[width=\textwidth]{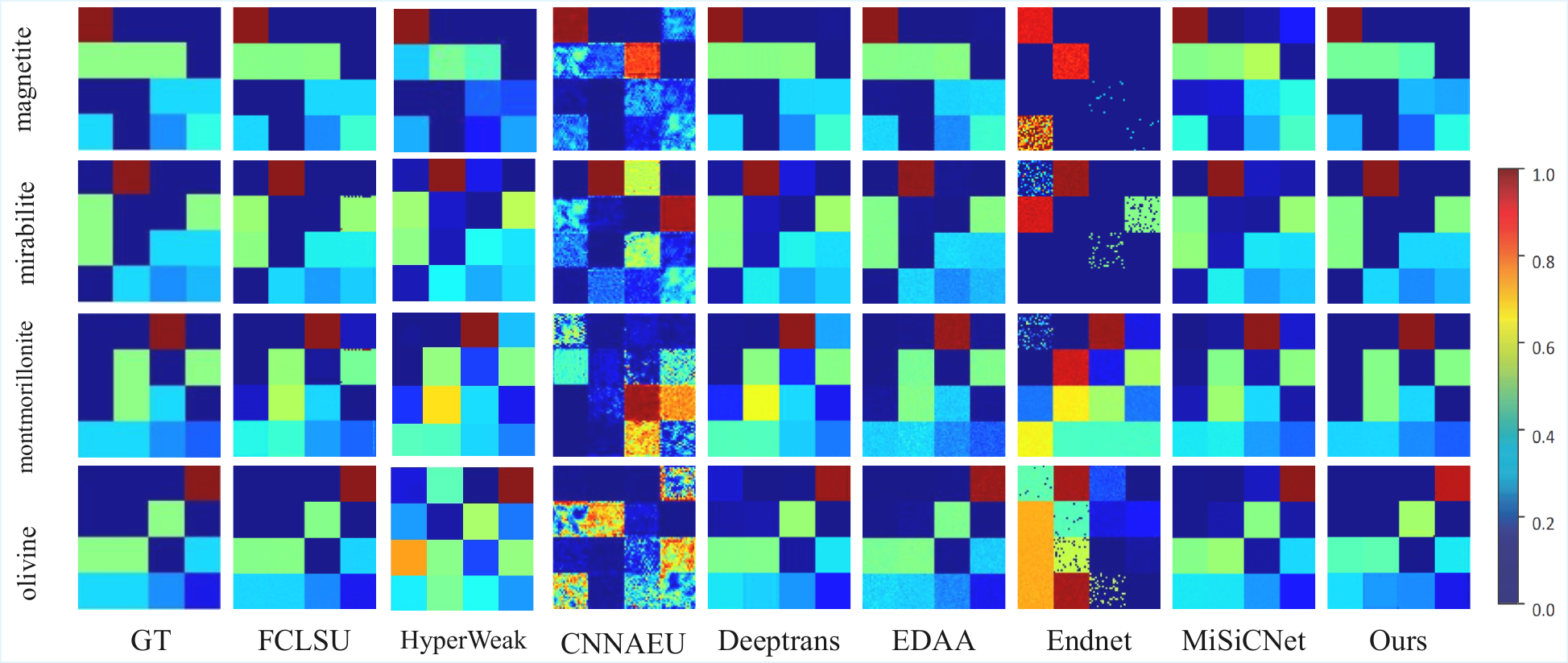}
    \caption{Abundance maps generated using the jet colourmap for the synthetic dataset. Each small image represents the spatial distribution of a specific endmember.}
    \label{fig:S1_abundance_maps}
\end{figure*}

\begin{figure*}[htbp]
    \centering
    \rotatebox{90}{\scriptsize{\hspace{0.8cm}Magnetite}}
    \includegraphics[width=0.128\linewidth]{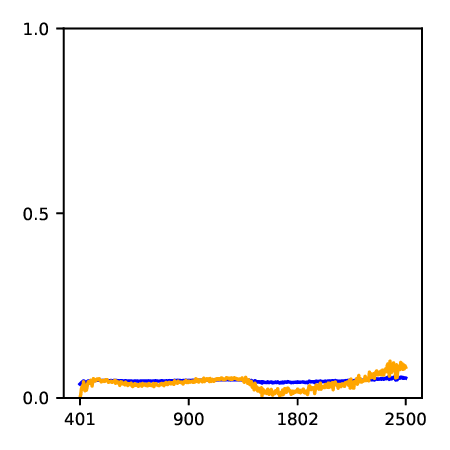}
    \hspace{-0.35cm}
    \includegraphics[width=0.128\linewidth]{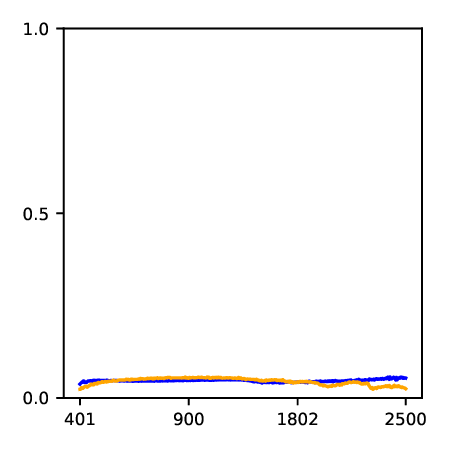}
    \hspace{-0.35cm}
    \includegraphics[width=0.128\linewidth]{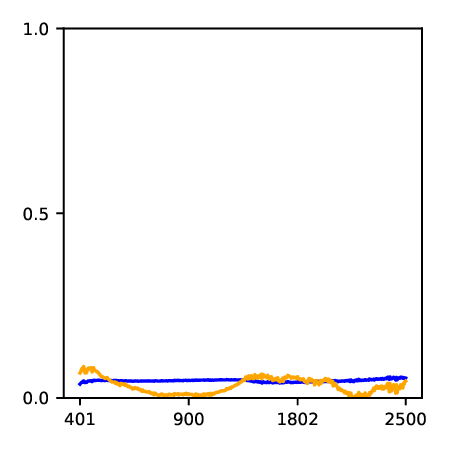}
    \hspace{-0.35cm}
    \includegraphics[width=0.128\linewidth]{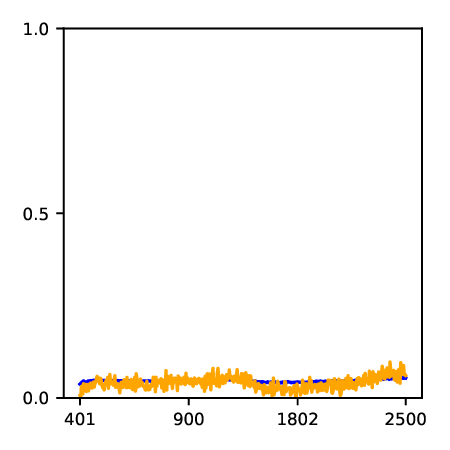}
    \hspace{-0.35cm}
    \includegraphics[width=0.128\linewidth]{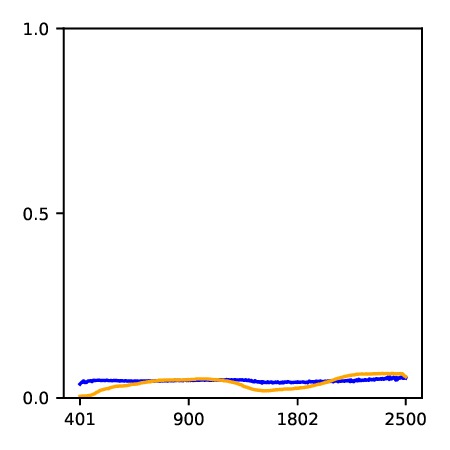}
    \hspace{-0.35cm}
    \includegraphics[width=0.128\linewidth]{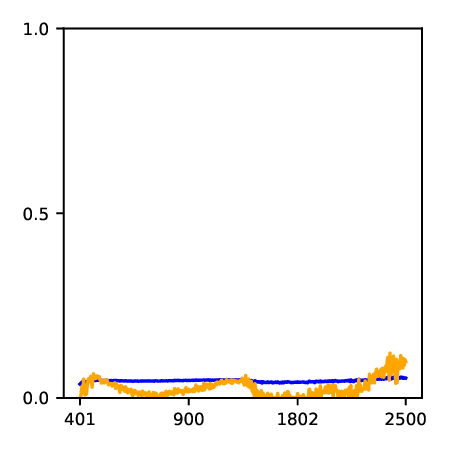}
    \hspace{-0.35cm}
    \includegraphics[width=0.128\linewidth]{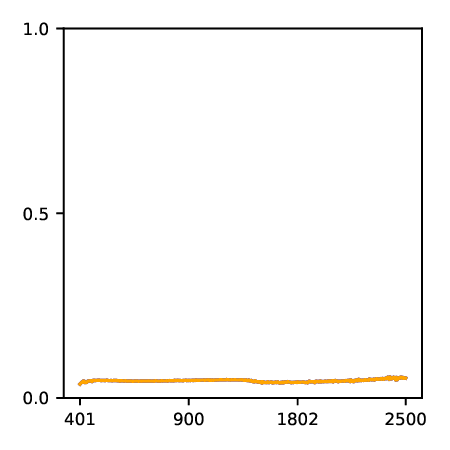}
    \hspace{-0.35cm}
    \includegraphics[width=0.128\linewidth]
    {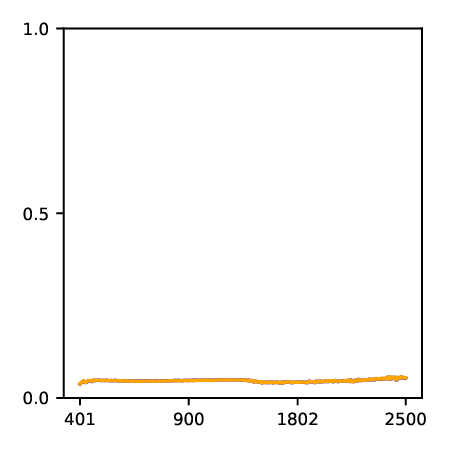}
\\
 \rotatebox{90}{\scriptsize{\hspace{0.6cm}Mirabilite}}
    \includegraphics[width=0.128\linewidth]{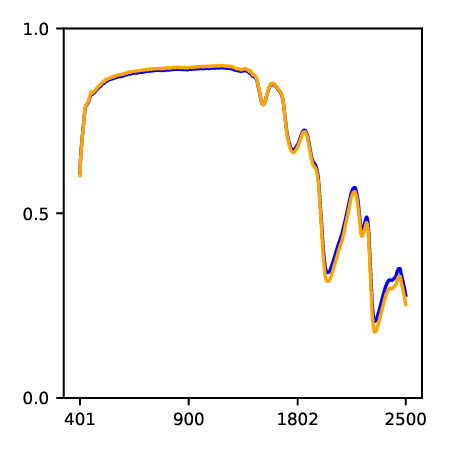}
    \hspace{-0.35cm}
    \includegraphics[width=0.128\linewidth]{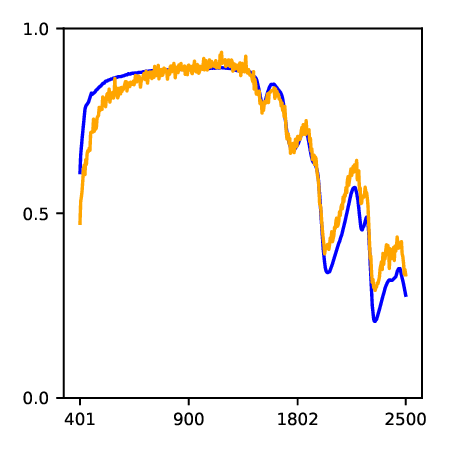}
    \hspace{-0.35cm}
    \includegraphics[width=0.128\linewidth]{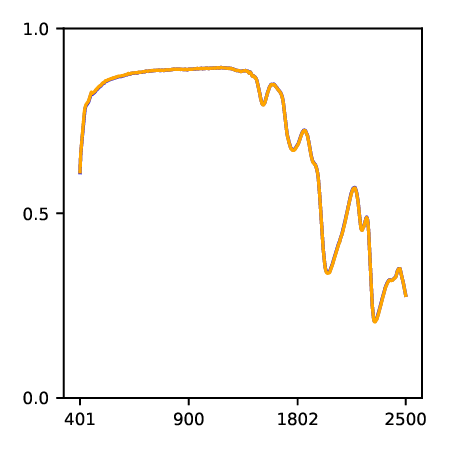}
    \hspace{-0.35cm}
    \includegraphics[width=0.128\linewidth]{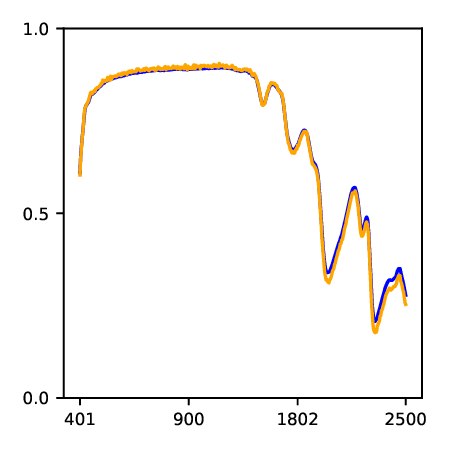}
    \hspace{-0.35cm}
    \includegraphics[width=0.128\linewidth]{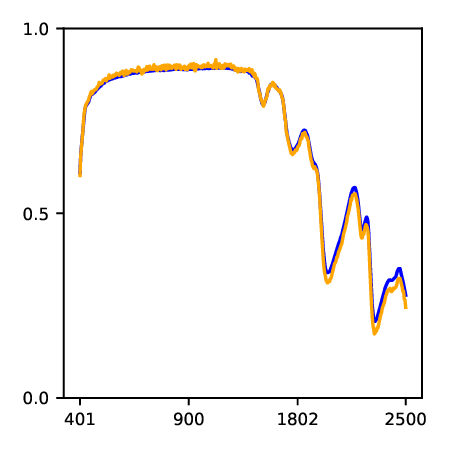}
    \hspace{-0.35cm}
    \includegraphics[width=0.128\linewidth]{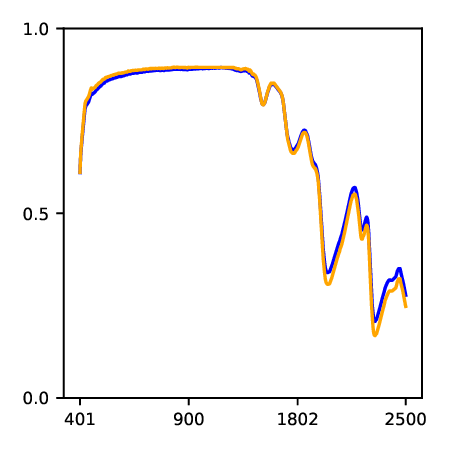}
    \hspace{-0.35cm}
    \includegraphics[width=0.128\linewidth]{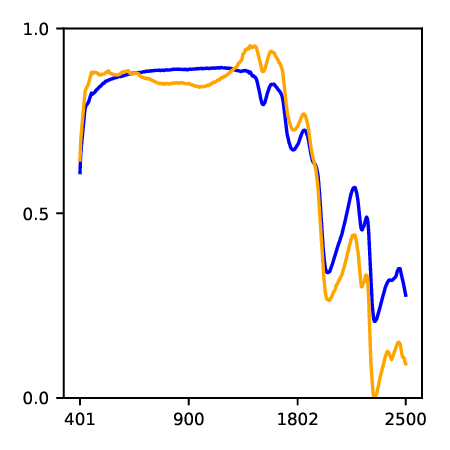}
    \hspace{-0.35cm}
    \includegraphics[width=0.128\linewidth]
    {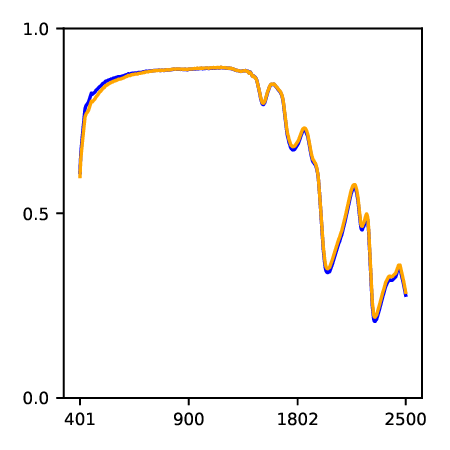}
\\
 \rotatebox{90}{\scriptsize{\hspace{0.6cm}Montmorillonite}}
    \includegraphics[width=0.128\linewidth]{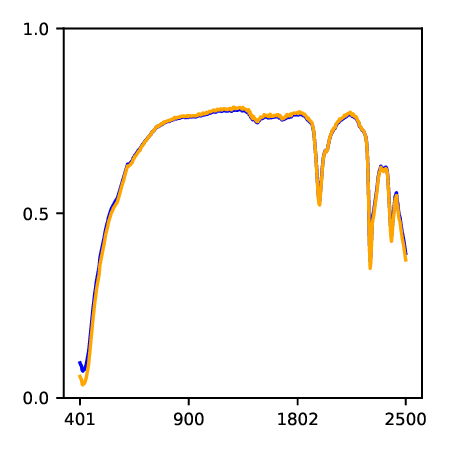}
    \hspace{-0.35cm}
    \includegraphics[width=0.128\linewidth]{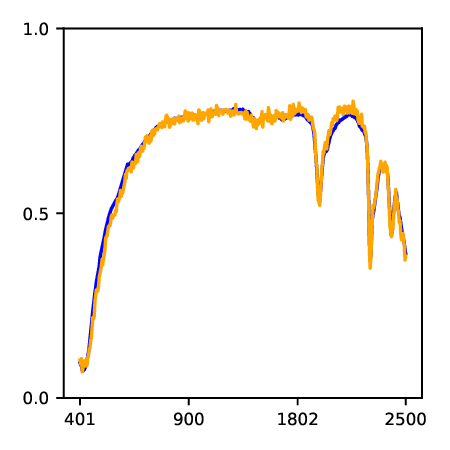}
    \hspace{-0.35cm}
    \includegraphics[width=0.128\linewidth]{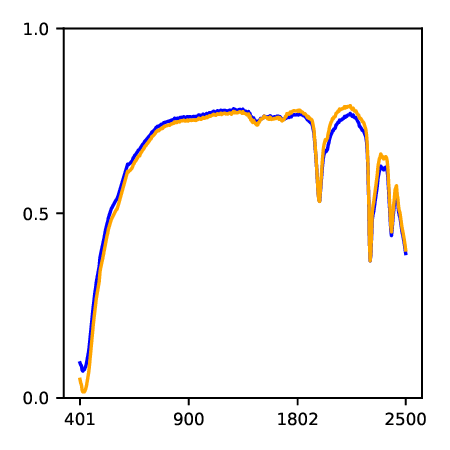}    \hspace{-0.35cm}
    \includegraphics[width=0.128\linewidth]{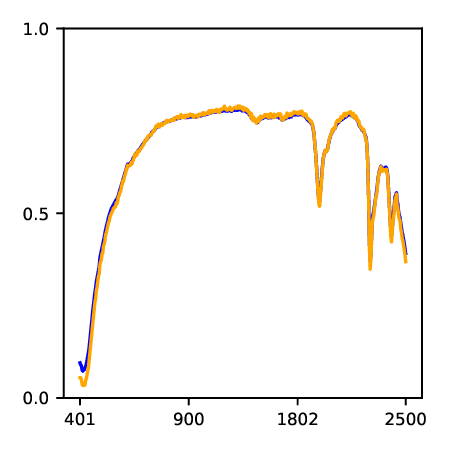}
    \hspace{-0.35cm}
    \includegraphics[width=0.128\linewidth]{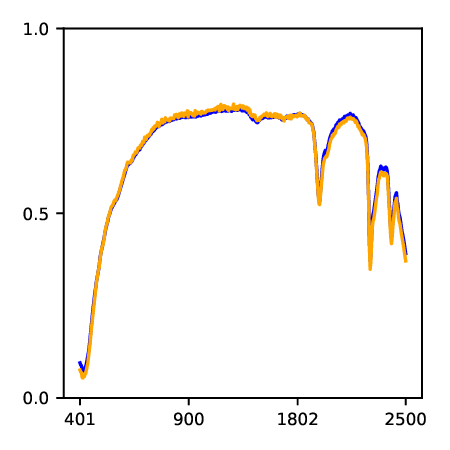}
    \hspace{-0.35cm}
    \includegraphics[width=0.128\linewidth]{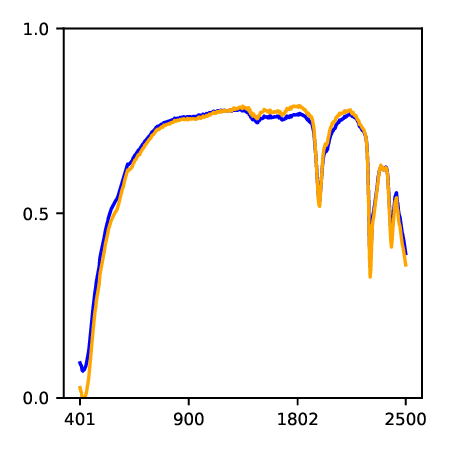}
    \hspace{-0.35cm}
    \includegraphics[width=0.128\linewidth]{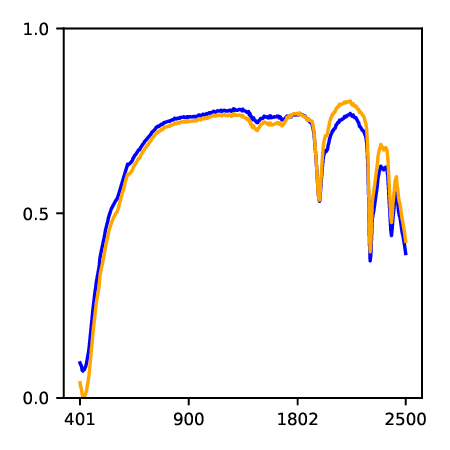}
    \hspace{-0.35cm}
    \includegraphics[width=0.128\linewidth]
    {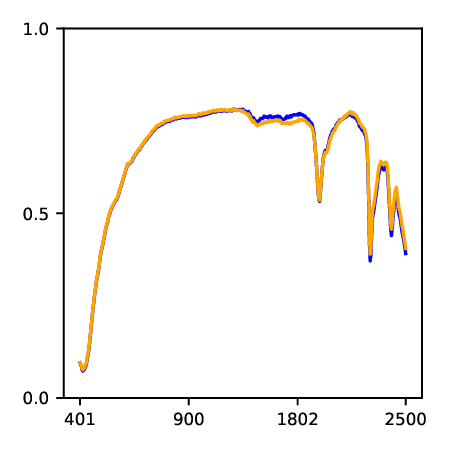}
\\
 \rotatebox{90}{\scriptsize{\hspace{0.9cm}Olivine}}
    \includegraphics[width=0.128\linewidth]{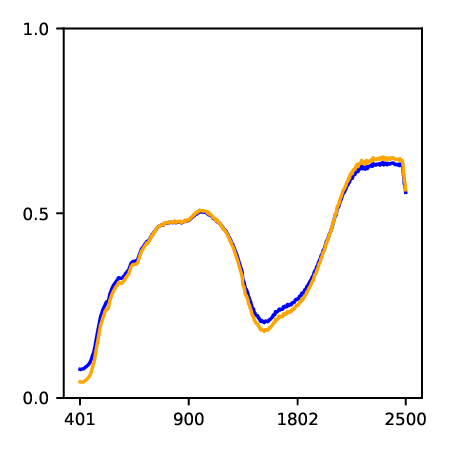}
    \hspace{-0.35cm}
    \includegraphics[width=0.128\linewidth]{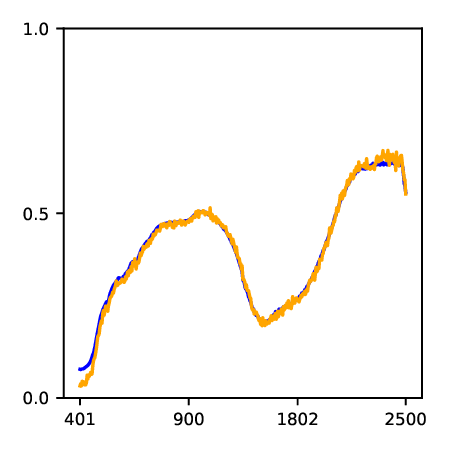}    \hspace{-0.35cm}
    \includegraphics[width=0.128\linewidth]{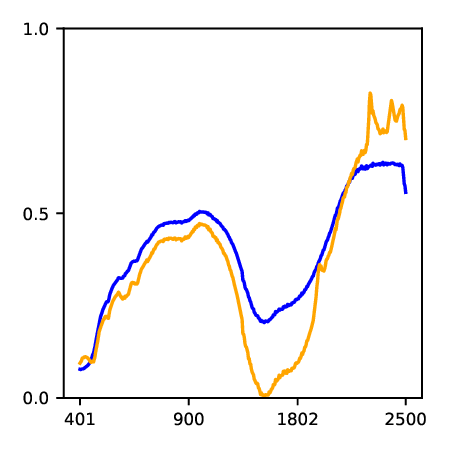}
    \hspace{-0.35cm}
    \includegraphics[width=0.128\linewidth]{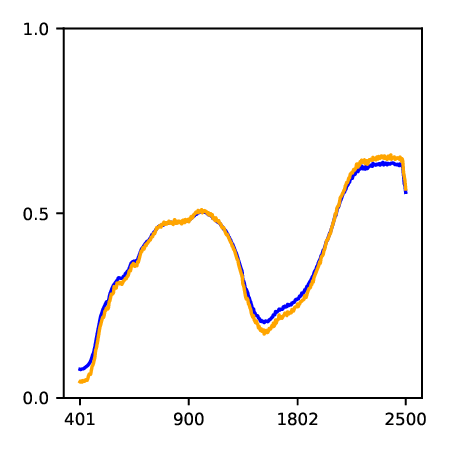}
    \hspace{-0.35cm}
    \includegraphics[width=0.128\linewidth]{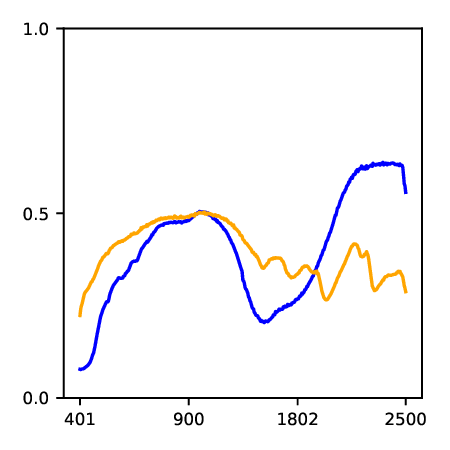}
    \hspace{-0.35cm}
    \includegraphics[width=0.128\linewidth]{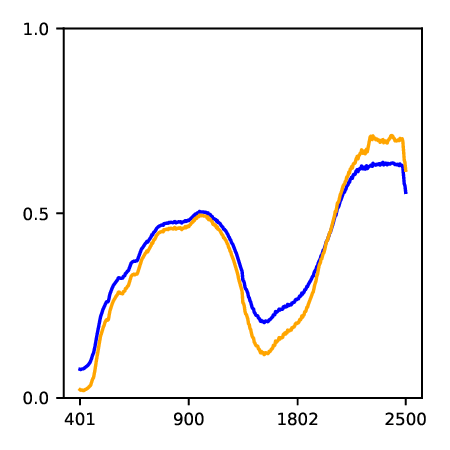}
    \hspace{-0.35cm}
    \includegraphics[width=0.128\linewidth]{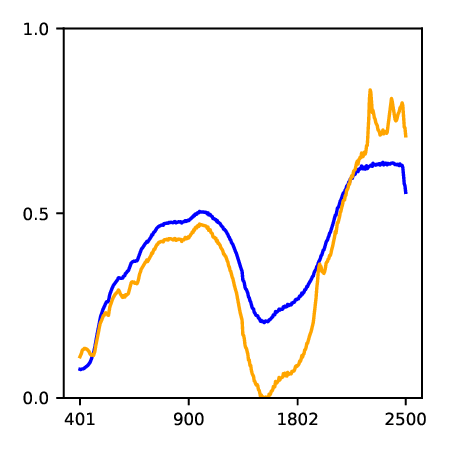}
    \hspace{-0.35cm}
    \includegraphics[width=0.128\linewidth]
    {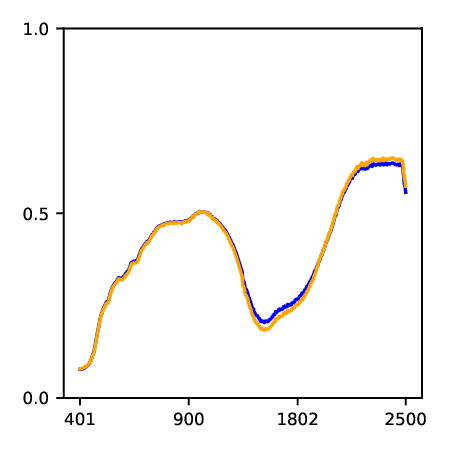}
    \begin{flushleft}
    \scriptsize{\hspace{0.5cm}FCLSU~\cite{heinz1999fully, nascimento2005vertex} \hspace{0.6cm} CNNAEU~\cite{palsson2020convolutional} \hspace{0.5cm} DeepTrans~\cite{9848995} \hspace{0.6cm} EDAA~\cite{zouaoui2023entropic} \hspace{0.8cm} EndNet ~\cite{ozkan2018endnet}\hspace{0.8cm} MiSiCNet~\cite{rasti2022misicnet} \hspace{0.6cm} Hyperweak~\cite{shen2022toward} \hspace{0.6cm} Ours}
    \end{flushleft}
    \caption{Visual comparison synthetic dataset of endmember spectra obtained by different unmixing techniques. The blue curves correspond to the ground‐truth spectra, while the orange curves denote the spectra estimated by each method.}
    \label{fig:end-S1}
\end{figure*}

\begin{table}[thbp]
\centering
\caption{Root Mean Square Error (RMSE) comparison on the simulated S1 dataset. The S1 dataset is constructed by linearly mixing four materials—magnetite, mirabilite, montmorillonite, and olivine—under a strict linear mixing model. The table reports RMSE values for each endmember and their average (RMSE\_Mean). Lower values indicate better abundance estimation accuracy. The best results are marked in red.}
\resizebox{\columnwidth}{!}{
\begin{tabular}{lccccc}
\textbf{Method} & Magnetite &Mirabilite & Montmorillonite & Olivine & \textbf{RMSE\_Mean} \\
\midrule
FCLSU~\cite{heinz1999fully, nascimento2005vertex}  & 0.0151 & 0.0315 & 0.0198 & 0.0146 & 0.0205\\
Hyperweak~\cite{shen2022toward} & 0.0379 & 0.0436 & 0.1145 & 0.1784 & 0.0936\\
CNNAEU~\cite{palsson2020convolutional}     & 0.0162 & 0.0506 & 0.0166 & 0.0130 & 0.0241 \\
DeepTrans~\cite{9848995}  & 0.0206 & 0.0153 & 0.0884 & 0.0137 & 0.0345 \\
EDAA~\cite{zouaoui2023entropic}       & 0.0155 & 0.0119 & 0.0106 & 0.0156 & 0.0134 \\
EndNet ~\cite{ozkan2018endnet}    & 0.0155 & 0.0140 & \textcolor{red}{\textbf{0.0099}} & 0.1526 & 0.0480 \\
MiSiCNet~\cite{rasti2022misicnet}   & 0.0287 & 0.0158 & 0.0226 & 0.0491 & 0.0290 \\
\rowcolor{gray!10}
Ours & \textcolor{red}{\textbf{ 0.0152}} & \textcolor{red}{\textbf{0.0150}} & 0.0109 & \textcolor{red}{\textbf{0.0114}} & \textcolor{red}{\textbf{0.0131}} \\
\bottomrule
\end{tabular}}
\label{tab:rmse_compare}
\end{table}

\begin{table}[thbp]
\centering
\scriptsize
\caption{Spectral Angle Distance (SAD) comparison on the simulated S1 dataset. SAD is used to evaluate the angular similarity between the estimated and ground truth abundance vectors. The S1 dataset consists of hyperspectral pixels with 459 bands ranging from 400 nm to 2500 nm and includes challenging ternary mixtures involving low-reflectance materials. Lower SAD values represent more accurate abundance direction estimation. The best results are marked in red.}
\resizebox{\columnwidth}{!}{
\begin{tabular}{lccccc}
\toprule
\textbf{Method} & Magnetite & Mirabilite & Montmorillonite & Olivine & \textbf{SAD\_Mean} \\
\midrule
FCLSU~\cite{heinz1999fully, nascimento2005vertex}  & 0.3300 & 0.0153 & 0.0343 & 0.1340 & 0.1284 \\ 
Hyperweak~\cite{shen2022toward} & \textcolor{red}{\textbf{0.0015}} & 0.1202 & 0.0457 & 0.2456 & 0.1033 \\
CNNAEU~\cite{palsson2020convolutional}     & 0.2194 & 0.0675 & 0.0244 & 0.0402 & 0.0879 \\
DeepTrans~\cite{9848995}  & 0.1281 & 0.0508 & 0.0342 & 0.0762 & 0.0723 \\
EDAA~\cite{zouaoui2023entropic}       & 0.3381 & 0.0159 & 0.0155 & 0.0362 & 0.1014 \\
EndNet ~\cite{ozkan2018endnet}    & 0.3392 & 0.0187 & \textcolor{red}{\textbf{0.0145}} & 0.3617 & 0.1835 \\
MiSiCNet~\cite{rasti2022misicnet}   & 0.6608 & 0.0211 & 0.0331 & 0.1141 & 0.2073 \\
\rowcolor{gray!10}
Ours      & 0.0262 & \textcolor{red}{\textbf{0.0084}} & 0.0240 & \textcolor{red}{\textbf{0.0296}} & \textcolor{red}{\textbf{0.0221}} \\
\bottomrule
\end{tabular}
}
\label{tab:sad_compare}
\end{table}

\subsection{Synthetic Experiment}
Simulated datasets are widely employed to validate the effectiveness of hyperspectral unmixing, with their primary advantage being the availability of precise endmember and abundance ground truths (GT), thereby providing a reliable benchmark for performance evaluation. 

To systematically assess the robustness of the proposed \emph{WS-Net} under noise interference, unmixing experiments were conducted under five different signal-to-noise ratio (SNR) conditions, namely 10 dB, 20 dB, 30 dB, 40 dB, and 50 dB. 
Fig.~\ref{fig:S1_abundance_maps} and Fig.~\ref{fig:end-S1} illustrate the abundance maps and endmember spectral comparisons obtained without noise. From a qualitative perspective, it can be observed that the abundance maps and spectral signatures produced by \emph{WS-Net} are highly consistent with the ground truth, with spatial distributions and spectral shapes superior to those of other competing methods. 

In terms of abundance estimation (Table~\ref{tab:rmse_compare}), \emph{WS-Net} achieves the lowest average RMSE of 0.0131, outperforming all baseline methods, including EDAA~\cite{zouaoui2023entropic}  and FCLSU~\cite{heinz1999fully, nascimento2005vertex}. Particularly for the weak signal endmember magnetite, \emph{WS-Net} obtains an RMSE of 0.0152, which is also the best among all methods. 

Regarding spectral angle similarity (Table~\ref{tab:sad_compare}), \emph{WS-Net} yields an average SAD of 0.0221, which is significantly lower than those of other methods, such as DeepTrans~\cite{9848995}  and FCLSU~\cite{heinz1999fully, nascimento2005vertex}. Although the SAD value for the weak-signal endmember Magnetite is slightly higher than that of certain methods, \emph{WS-Net} demonstrates a superior overall balance and consistent advantages in average performance.

These quantitative results further confirm that, compared with other algorithms, \emph{WS-Net} achieves more accurate and stable unmixing performance, particularly in terms of overall accuracy and the modelling of weak-signal endmembers.

\subsection{Evaluation Under Degraded Signal Conditions}
\begin{figure}[htbp]
    \centering
    \includegraphics[width=0.49\linewidth]{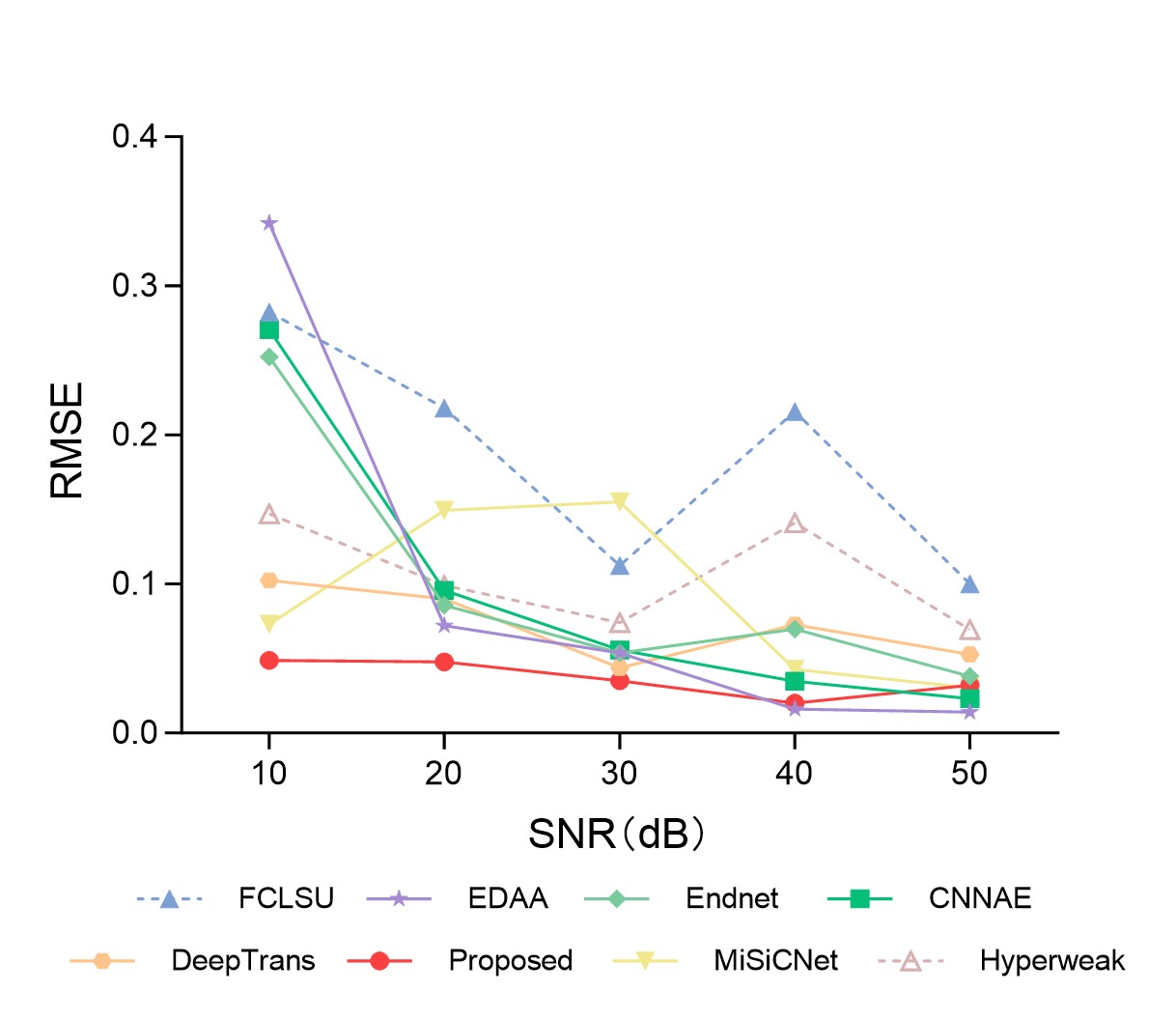}
    \includegraphics[width=0.49\linewidth]{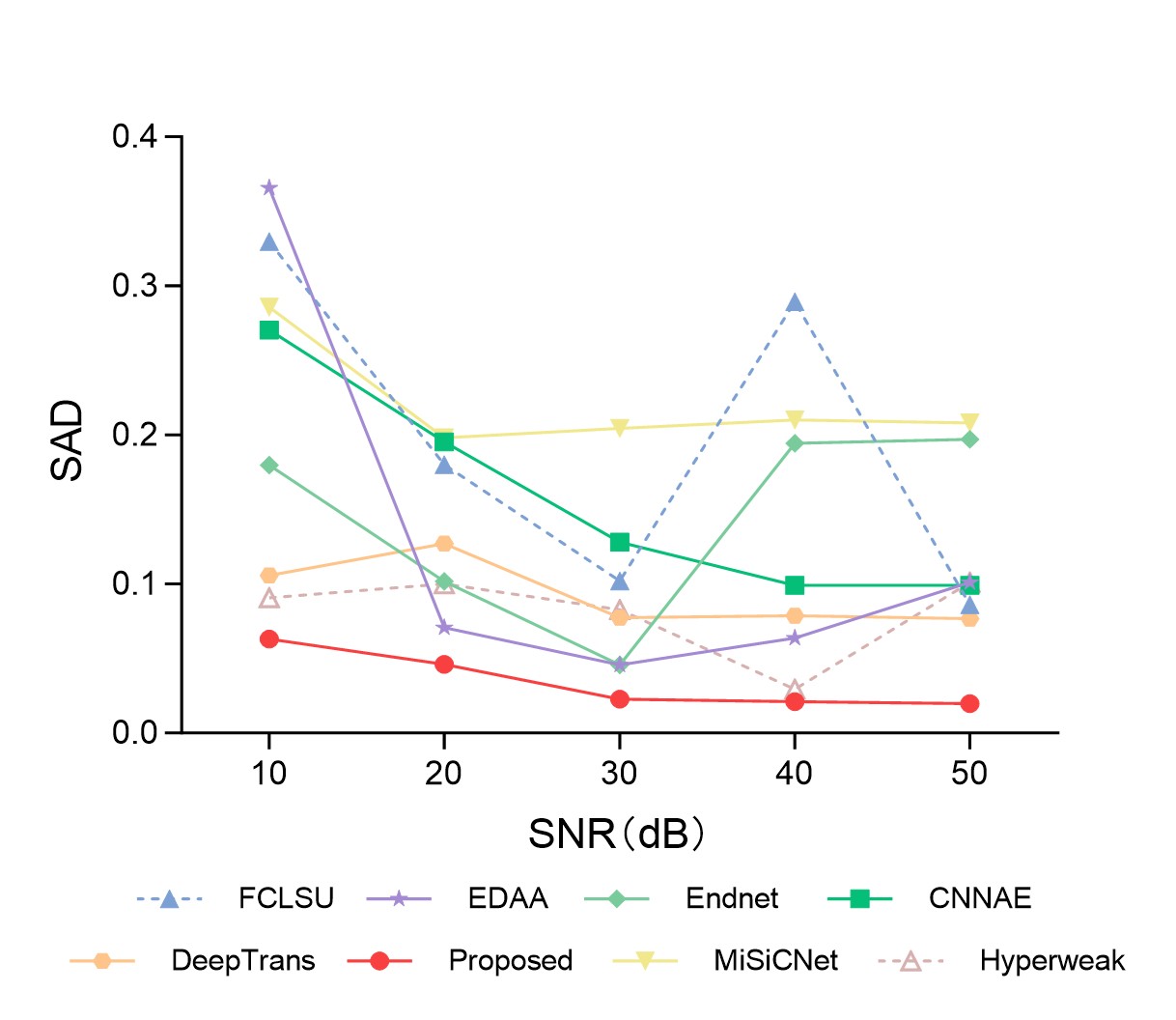}\\
    \includegraphics[width=0.49\linewidth]{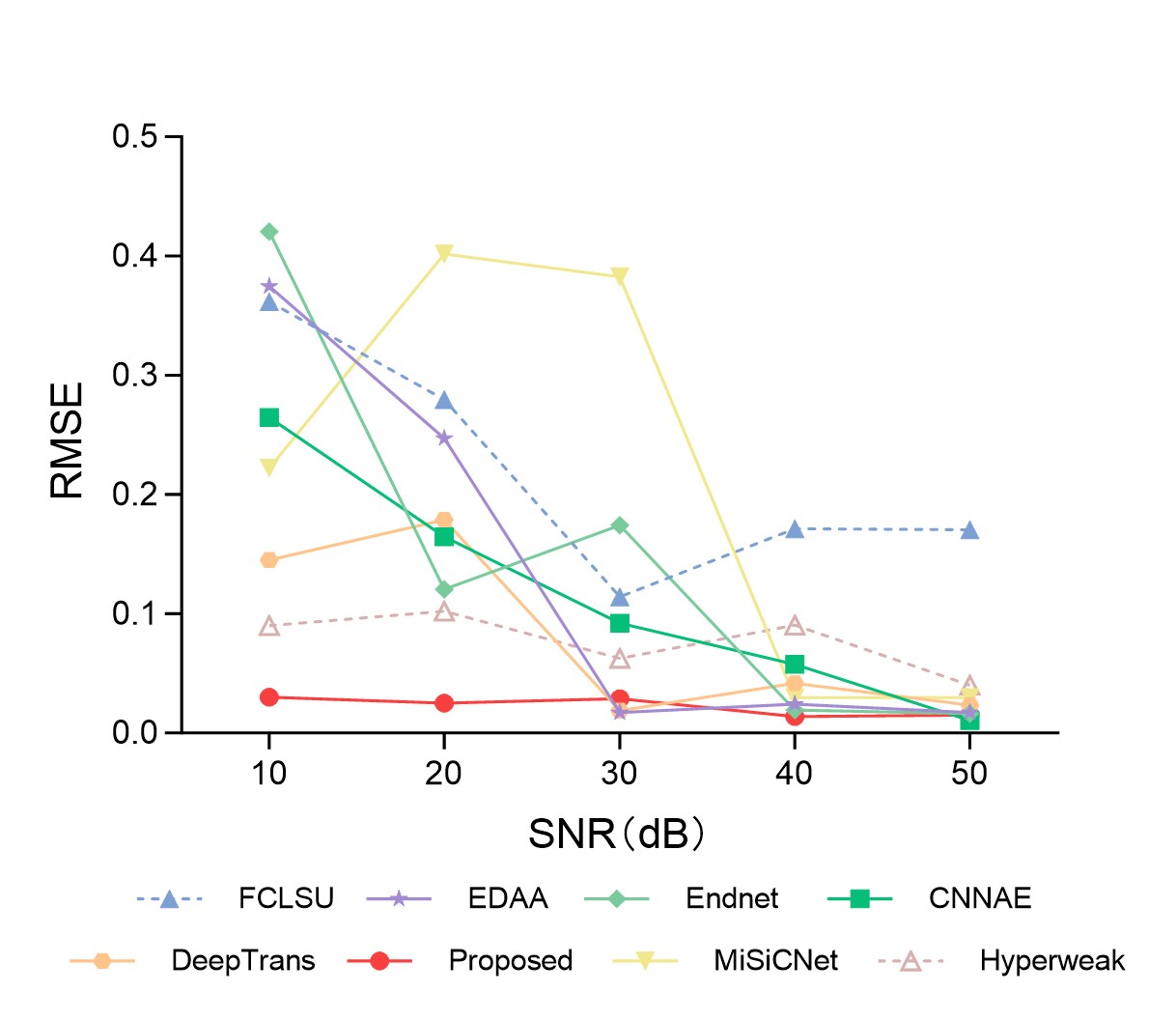}
    \includegraphics[width=0.49\linewidth]{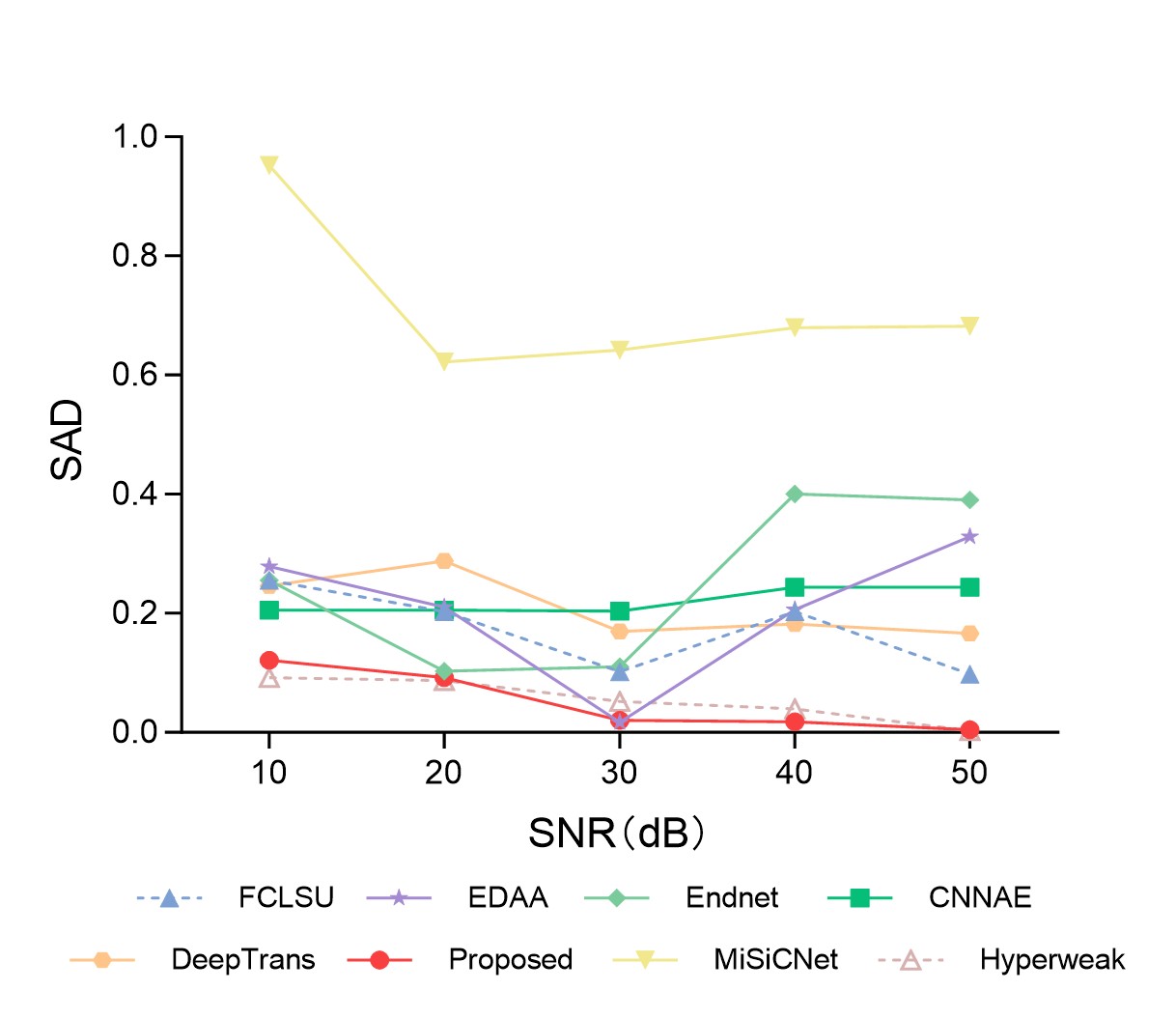}

    \caption{Performance comparison of different unmixing algorithms in terms of average and material-specific metrics. 
    (a) Average RMSE across all materials. 
    (b) Average SAD across all materials. 
    (c) weak signal RMSE for Magnetite. 
    (d) weak signal  SAD for Magnetite. 
    Lower values indicate better unmixing accuracy.}
    \label{fig:performance_curves}
\end{figure}
Fig.~\ref{fig:performance_curves} presents the performance comparison of different unmixing methods under various signal-to-noise ratio (SNR) conditions, where the evaluation metrics include the average RMSE, SAD, and the unmixing accuracy of the magnetite endmember. The results clearly demonstrate that the proposed \emph{WS-Net} consistently achieves the lowest RMSE and SAD values across all noise levels. Moreover, its performance curves remain smooth with negligible fluctuations, highlighting its outstanding robustness. In contrast, other algorithms exhibit evident performance degradation under low-SNR conditions (e.g., 10 dB and 20 dB), with significantly larger variations in their curves. 

Further analysis of the weak-signal endmember magnetite shows that the RMSE and SAD values of \emph{WS-Net} remain nearly close to zero across all SNR levels, whereas the errors of competing methods increase substantially under low-SNR scenarios. This indicates that \emph{WS-Net} is capable of stably and accurately recovering weak-signal endmembers even in complex mixtures and high-noise environments, underscoring its superiority in weak-signal modelling. 

In summary, the systematic experimental results on simulated data confirm that \emph{WS-Net} significantly outperforms existing algorithms in both overall accuracy and weak-signal endmember modelling. Regardless of noise level, its RMSE and SAD curves exhibit remarkable stability, fully validating its effectiveness and robustness for practical hyperspectral unmixing tasks.

\subsection{Samson Real Dataset Experiment}
\begin{figure*}[t]
    \centering
    \includegraphics[width=\textwidth]{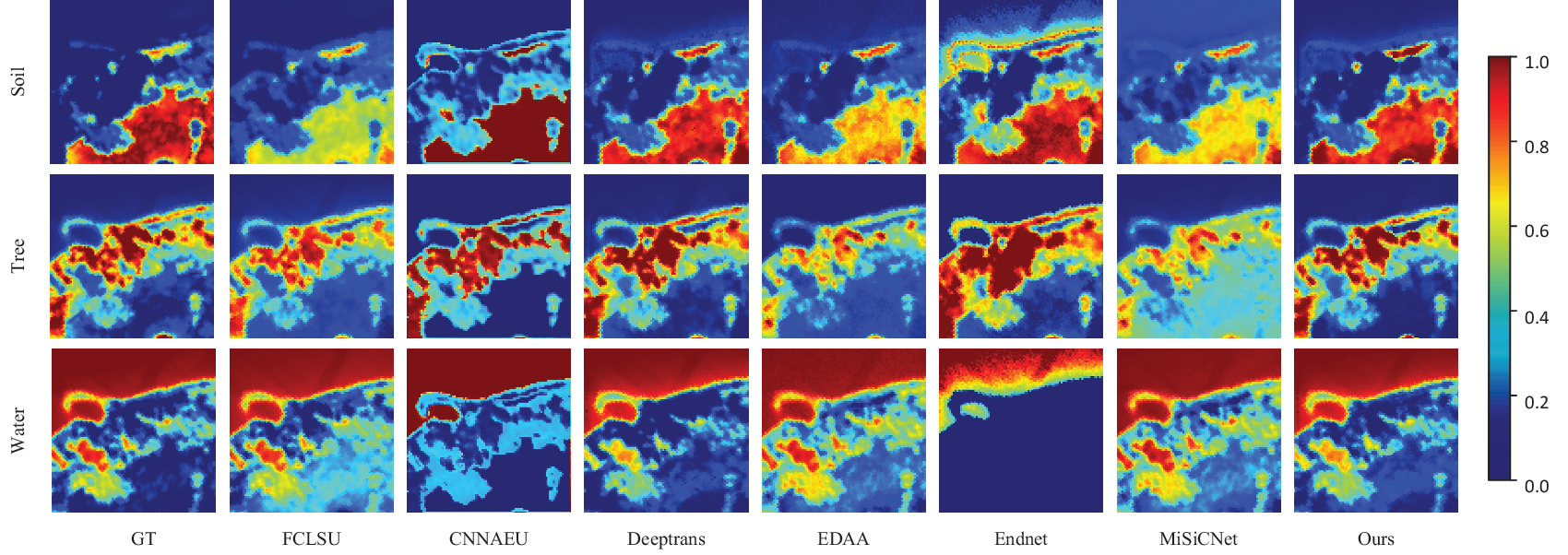}
    \caption{Abundance maps generated using the jet colourmap for the Samson dataset. Each small image represents the spatial distribution of a specific endmember.}
    \label{fig:samson_abundance_maps}
\end{figure*}

\begin{figure*}[htbp]
    \centering
    \rotatebox{90}{\scriptsize{\hspace{1cm}Soil}}
    \includegraphics[width=0.128\linewidth]{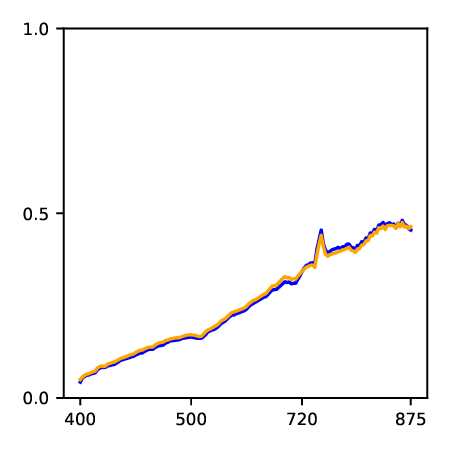}
    \hspace{-0.35cm}
    \includegraphics[width=0.128\linewidth]{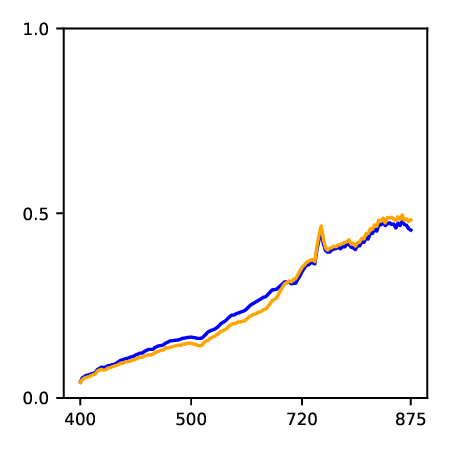}
    \hspace{-0.35cm}
    \includegraphics[width=0.128\linewidth]{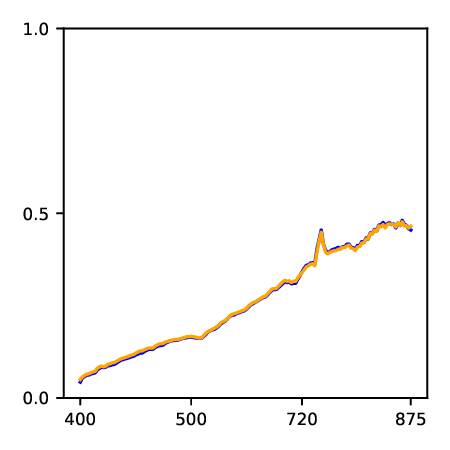}
    \hspace{-0.35cm}
    \includegraphics[width=0.128\linewidth]{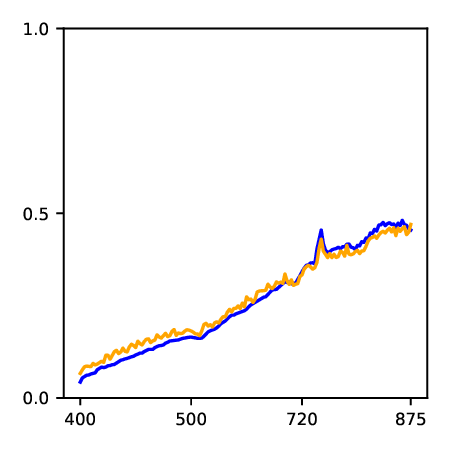}
    \hspace{-0.35cm}
    \includegraphics[width=0.128\linewidth]{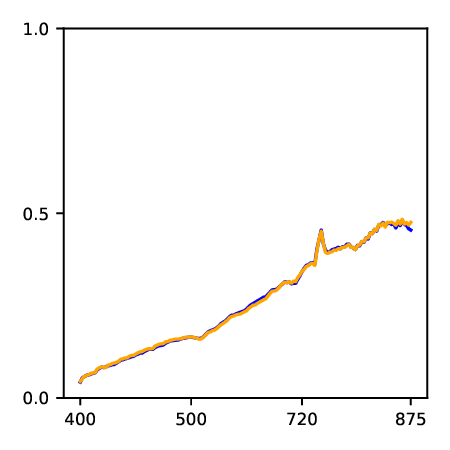}
    \hspace{-0.35cm}
    \includegraphics[width=0.128\linewidth]{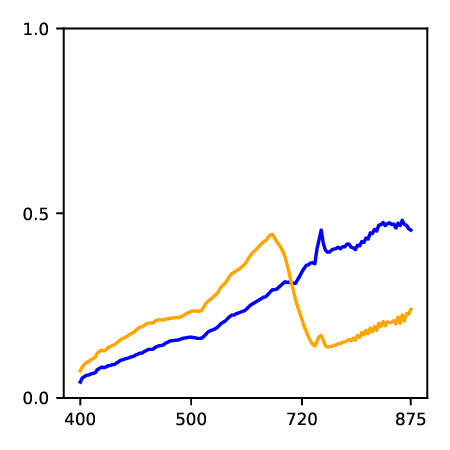}
    \hspace{-0.35cm}
    \includegraphics[width=0.128\linewidth]{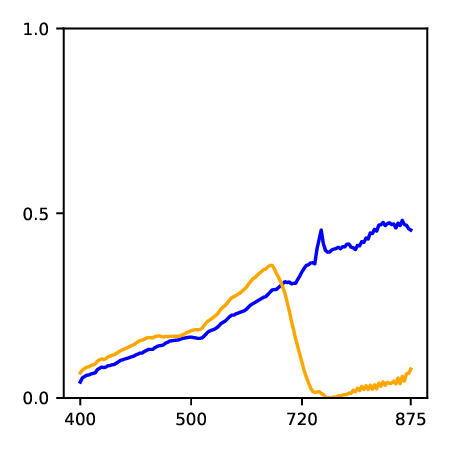}
    \hspace{-0.35cm}
    \includegraphics[width=0.128\linewidth]
    {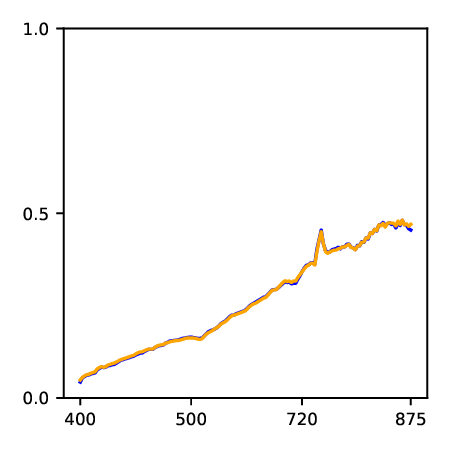}
\\
 \rotatebox{90}{\scriptsize{\hspace{1cm}Tree}}
    \includegraphics[width=0.128\linewidth]{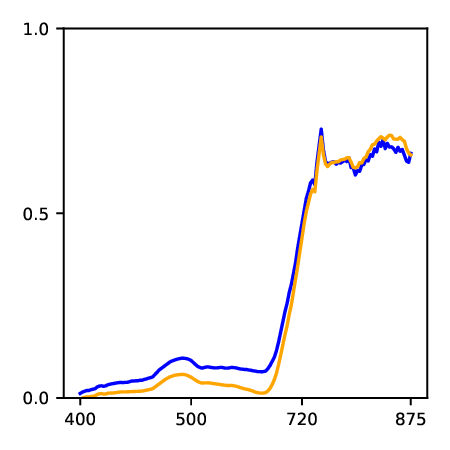}
    \hspace{-0.35cm}
    \includegraphics[width=0.128\linewidth]{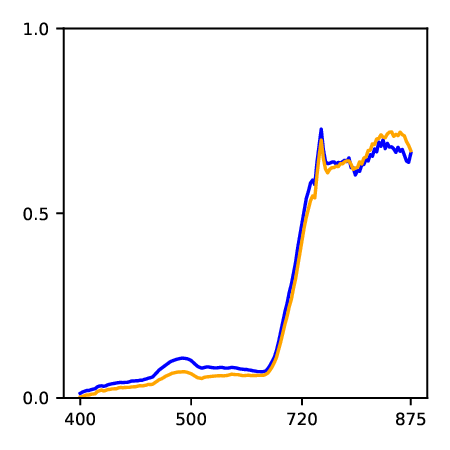}
    \hspace{-0.35cm}
    \includegraphics[width=0.128\linewidth]{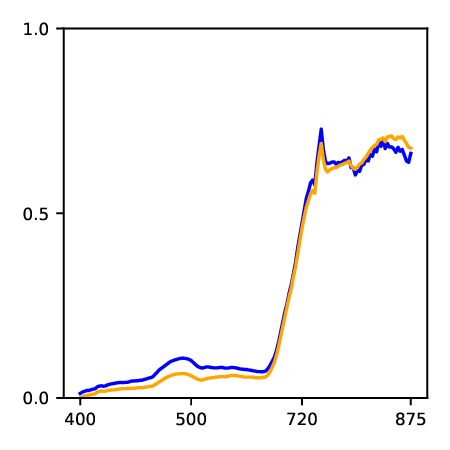}
    \hspace{-0.35cm}
    \includegraphics[width=0.128\linewidth]{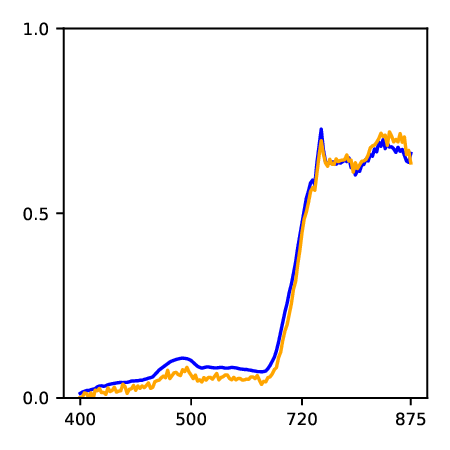}
    \hspace{-0.35cm}
    \includegraphics[width=0.128\linewidth]{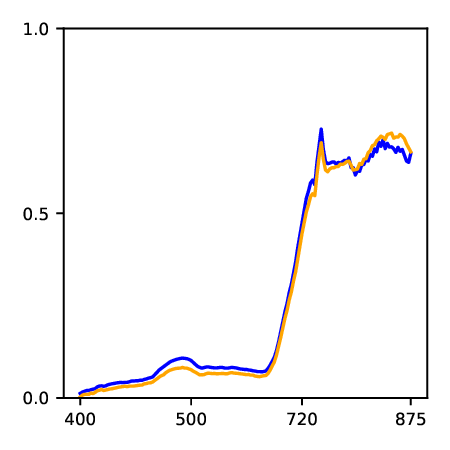}
    \hspace{-0.35cm}
    \includegraphics[width=0.128\linewidth]{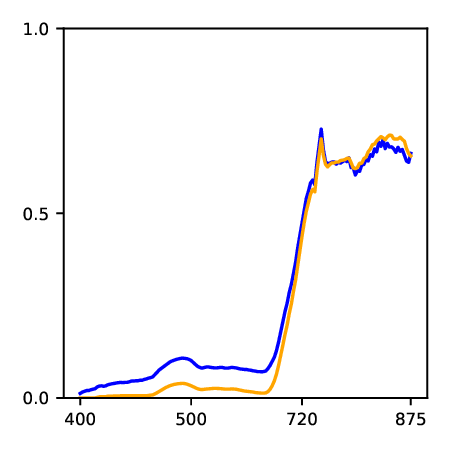}
    \hspace{-0.35cm}
    \includegraphics[width=0.128\linewidth]{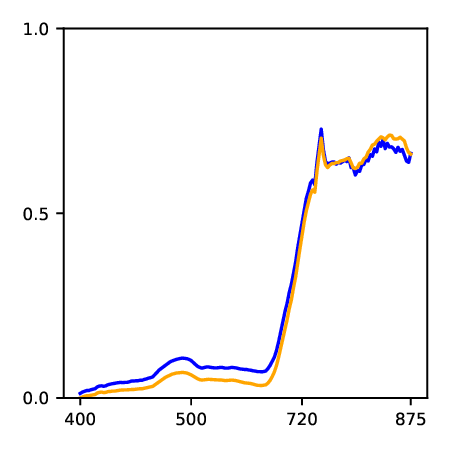}
    \hspace{-0.35cm}
    \includegraphics[width=0.128\linewidth]
    {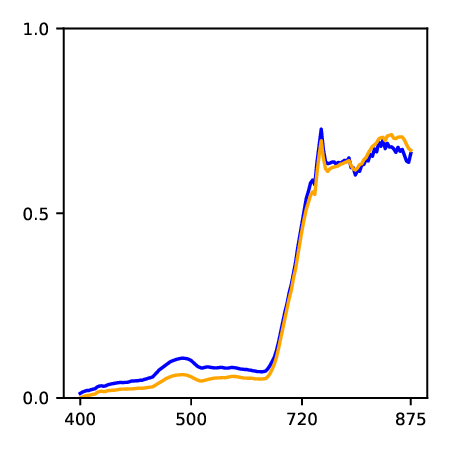}
\\
 \rotatebox{90}{\scriptsize{\hspace{1cm}Water}}
    \includegraphics[width=0.128\linewidth]{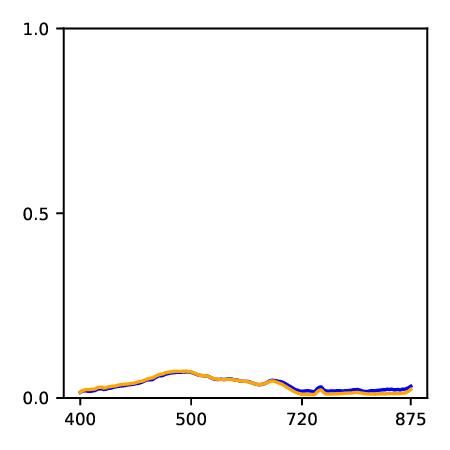}
    \hspace{-0.35cm}
    \includegraphics[width=0.128\linewidth]{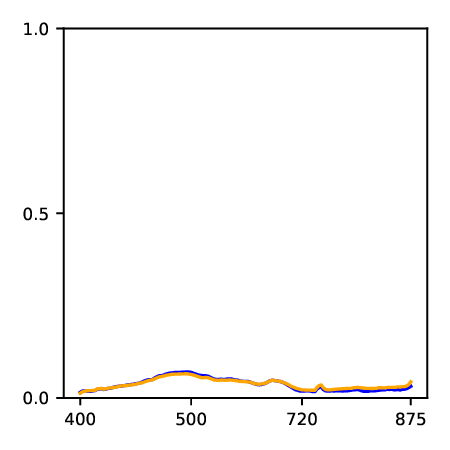}
    \hspace{-0.35cm}
    \includegraphics[width=0.128\linewidth]{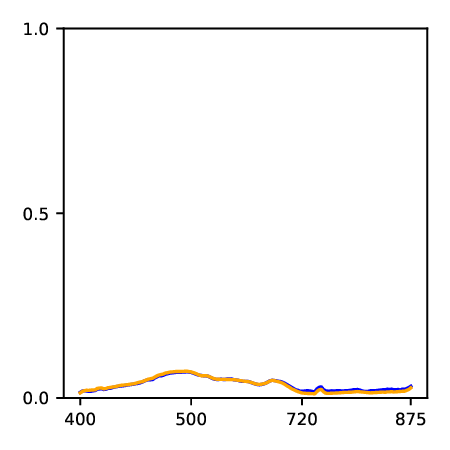}    \hspace{-0.35cm}
    \includegraphics[width=0.128\linewidth]{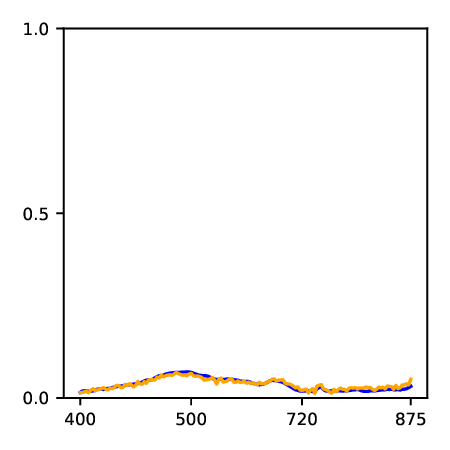}
    \hspace{-0.35cm}
    \includegraphics[width=0.128\linewidth]{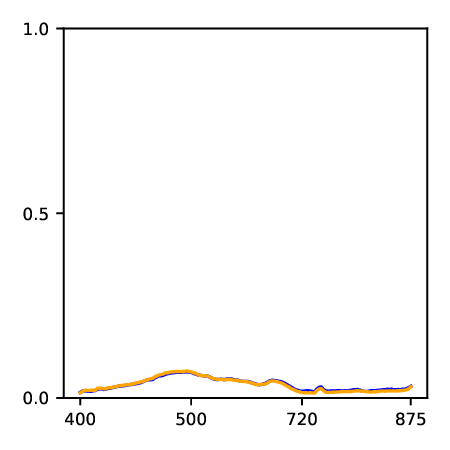}
    \hspace{-0.35cm}
    \includegraphics[width=0.128\linewidth]{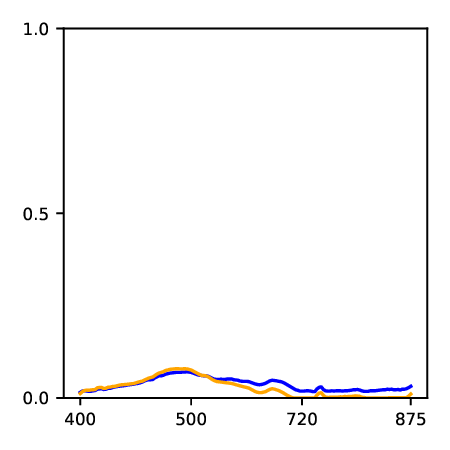}
    \hspace{-0.35cm}
    \includegraphics[width=0.128\linewidth]{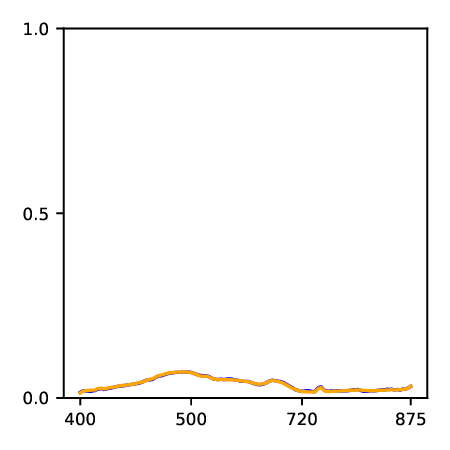}
    \hspace{-0.35cm}
    \includegraphics[width=0.128\linewidth]
    {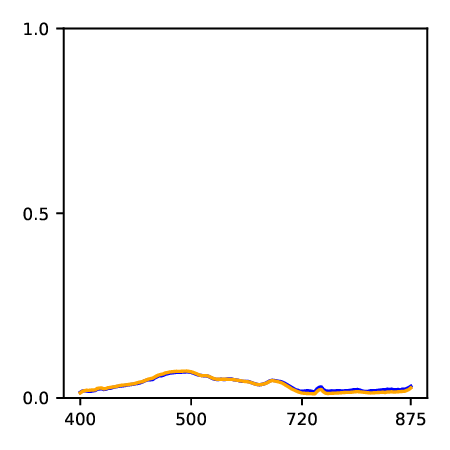}
    \begin{flushleft}
    \scriptsize{\hspace{0.5cm}FCLSU~\cite{heinz1999fully, nascimento2005vertex} \hspace{0.6cm} CNNAEU~\cite{palsson2020convolutional} \hspace{0.5cm} DeepTrans~\cite{9848995} \hspace{0.6cm} EDAA~\cite{zouaoui2023entropic} \hspace{0.8cm} EndNet ~\cite{ozkan2018endnet}\hspace{0.8cm} MiSiCNet~\cite{rasti2022misicnet} \hspace{0.6cm} Hyperweak~\cite{shen2022toward} \hspace{0.6cm} Ours}
    \end{flushleft}
    \caption{Visual comparison Samson dataset of endmember spectra obtained by different unmixing techniques. The blue curves correspond to the ground‐truth spectra, while the orange curves denote the spectra estimated by each method.}
    \label{fig:samson_end}
\end{figure*}

\begin{table}[t]
\centering
\scriptsize
\caption{Root Mean Square Error (RMSE) comparison on the Samson dataset. RMSE is computed per endmember and averaged across three endmembers. Lower values indicate better estimation performance. The best results are shown in red.}
\begin{tabular}{@{}lcccc@{}}
\toprule
\textbf{Method} & Soil & Tree & Water & \textbf{RMSE\_Mean} \\
\midrule
FCLSU~\cite{heinz1999fully, nascimento2005vertex}      & 0.1076 & 0.0371 & 0.0158 & 0.0535 \\
Hyperweak~\cite{shen2022toward} & 0.1576 & 0.2202 & 0.3589 & 0.2456\\
CNNAEU~\cite{palsson2020convolutional}     & 0.0164 & 0.0261 & 0.1040 & 0.0488 \\
DeepTrans~\cite{9848995}  & 0.0712 & 0.0683 & 0.0930 & 0.0775 \\
EDAA~\cite{zouaoui2023entropic}       & 0.0197 & 0.0278 & 0.0656 & 0.0377 \\
EndNet ~\cite{ozkan2018endnet}    & \textcolor{red}{\textbf{0.0036}} & 0.0263 & \textcolor{red}{\textbf{0.0145}} & \textcolor{red}{\textbf{0.0148}} \\
MiSiCNet~\cite{rasti2022misicnet}   & 0.1608 & 0.0449 & 0.0153 & 0.0737 \\
\rowcolor{gray!10}
Ours       & 0.0702 & \textcolor{red}{\textbf{0.0249}} & 0.0152 & 0.0368 \\
\bottomrule
\end{tabular}
\label{tab:samson_rmse}
\end{table}

\begin{table}[t]
\centering
\scriptsize
\caption{Spectral Angle Distance (SAD) comparison on the Samson dataset. SAD measures angular similarity between predicted and ground truth abundance vectors. Lower values indicate better directional estimation. Best results are marked in red.}
\begin{tabular}{@{}lcccc@{}}
\toprule
\textbf{Method} & Soil & Tree & Water & \textbf{SAD\_Mean} \\
\midrule
FCLSU~\cite{heinz1999fully, nascimento2005vertex}      & 0.0260 & 0.0970 & 0.1461 & 0.0897 \\
Hyperweak~\cite{shen2022toward} & 0.9229 & 0.0758 & \textcolor{red}{\textbf{0.0282}} & 0.3423\\
CNNAEU~\cite{palsson2020convolutional}     & 0.0559 & 0.0683 & 0.1012 & 0.0752 \\
DeepTrans~\cite{9848995}  & 0.0120 & 0.0616 & 0.0898 & 0.0545 \\
EDAA~\cite{zouaoui2023entropic}       & 0.0674 & 0.0726 & 0.1400 & 0.0933 \\
EndNet ~\cite{ozkan2018endnet}    & 0.0124 & 0.0671 & 0.0619 & 0.0460 \\
MiSiCNet~\cite{rasti2022misicnet}   & 0.5810 & 0.1175 & 0.3924 & 0.3636 \\
\rowcolor{gray!10}
Ours       & \textcolor{red}{\textbf{0.0094}} &\textcolor{red}{\textbf{ 0.0604}} & 0.0436 & \textcolor{red}{\textbf{0.0378}} \\
\bottomrule
\end{tabular}
\label{tab:samson_sad}
\end{table}

To further validate the generalisation capability and real-world applicability of the proposed method, experiments were conducted on the Samson dataset. This dataset contains three primary land-cover types: Soil, Tree, and Water. Figs~\ref{fig:samson_abundance_maps} and~\ref{fig:samson_end} illustrate the abundance maps and endmember spectra obtained by different unmixing algorithms. 

From a qualitative perspective, the abundance maps produced by the proposed \emph{WS-Net} exhibit a closer alignment with the actual land-cover distribution, indicating that our method can more accurately identify and separate mixed pixels into their constituent classes. 

Quantitative evaluations are summarised in Tables~\ref{tab:samson_rmse} and \ref{tab:samson_sad}. In terms of the key metric SAD, our method achieves an average value of 0.0378, which is significantly lower than all competing approaches. This demonstrates that \emph{WS-Net} provides more precise estimation of endmember spectra. Although the average RMSE of our method is slightly higher than that of EndNet~\cite{ozkan2018endnet} and a few other baselines, the substantial advantage in SAD, along with superior RMSE performance on specific endmembers such as Soil and Water, highlights the enhanced spectral fidelity of our unmixing results. 

It is particularly noteworthy that \emph{WS-Net} exhibits strong capability in handling weak-signal endmembers. Our method achieves the lowest SAD and RMSE values for this endmember, further validating the effectiveness of the \emph{WS-Net} architecture, which is specifically designed to address weak-signal modeling challenges.

The evaluation on the Samson dataset further demonstrates the generalization capability of our method, particularly its superior performance in modeling weak-signal endmembers. Overall, these findings provide strong evidence that \emph{WS-Net} possesses remarkable advantages and reliability in hyperspectral unmixing tasks involving weak signals and highly mixed pixels, making it a promising and practical solution for real-world applications.

\subsection{Apex Real Dataset Experiment}
\begin{figure*}[htbp]
    \centering
    \includegraphics[width=\textwidth]{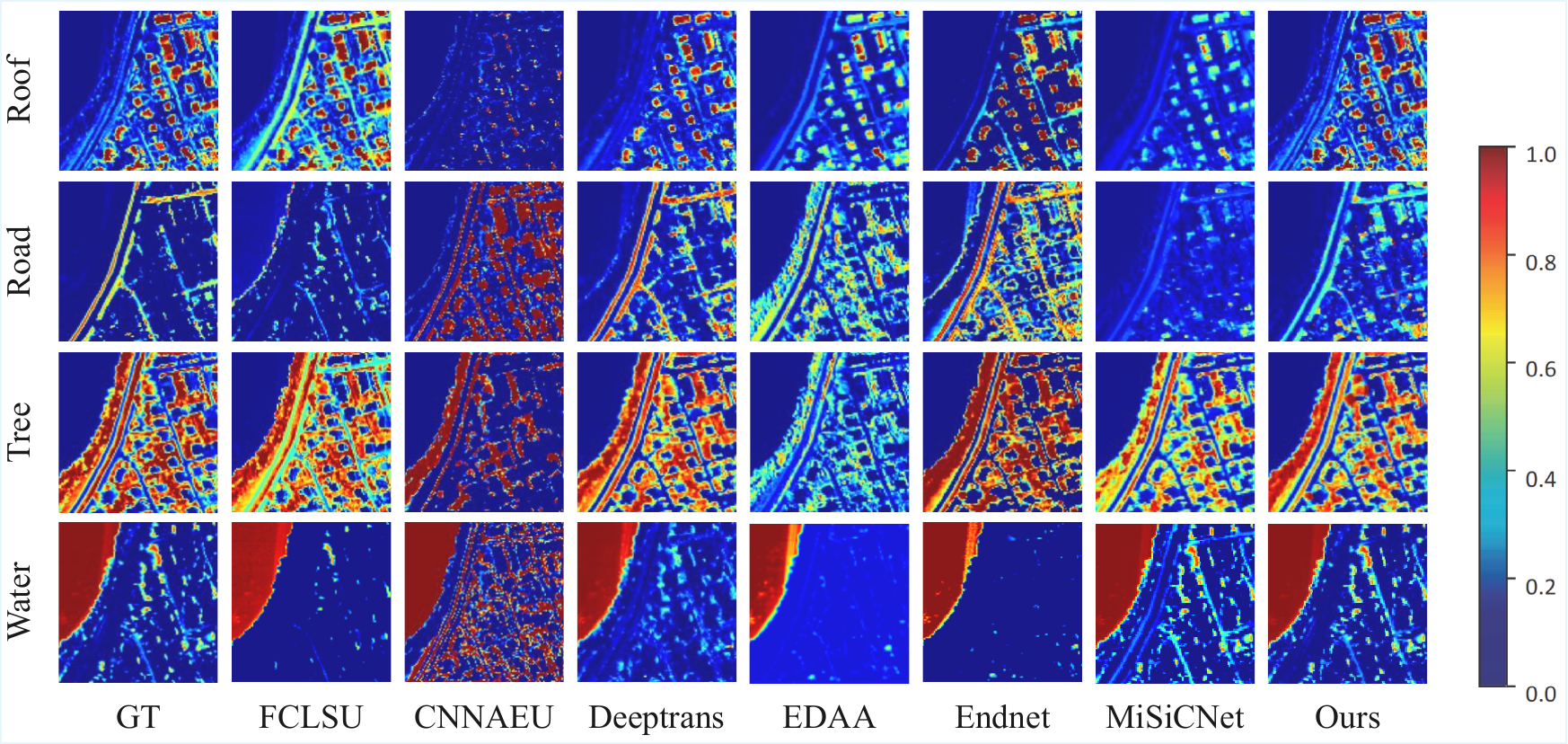}
    \caption{Abundance maps generated using the jet colormap for the apex dataset. Each small image represents the spatial distribution of a specific endmember.}
    \label{fig:apex_abundance_maps}
\end{figure*}
\begin{figure*}[htbp]
    \centering
    \rotatebox{90}{\scriptsize{\hspace{1cm}Roof}}
    \includegraphics[width=0.128\linewidth]{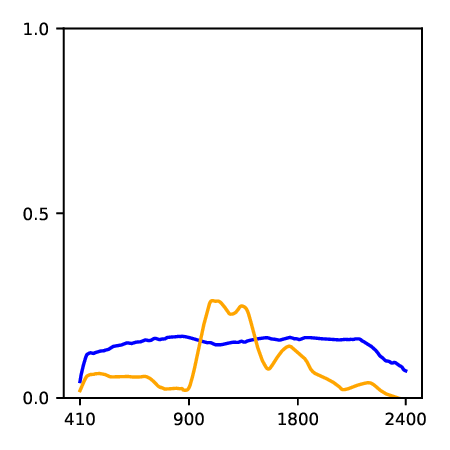}
    \hspace{-0.35cm}
    \includegraphics[width=0.128\linewidth]{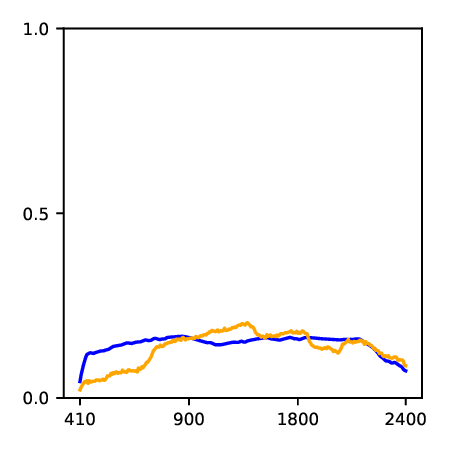}
    \hspace{-0.35cm}
    \includegraphics[width=0.128\linewidth]{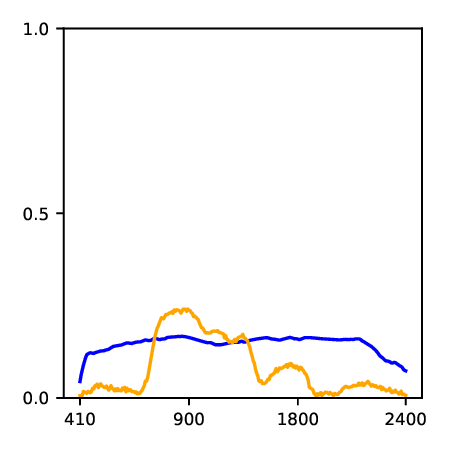}
    \hspace{-0.35cm}
    \includegraphics[width=0.128\linewidth]{fig/apex/APEX_Part_EDAA_EM1.eps}
    \hspace{-0.35cm}
    \includegraphics[width=0.128\linewidth]{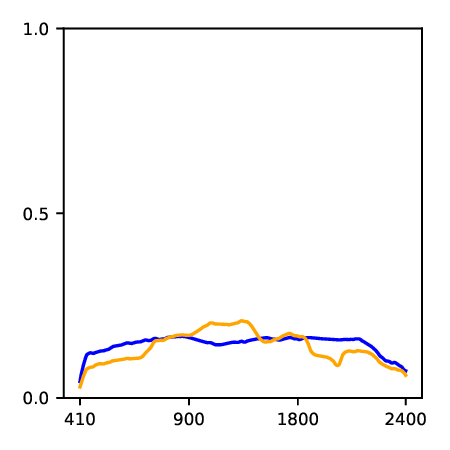}
    \hspace{-0.35cm}
    \includegraphics[width=0.128\linewidth]{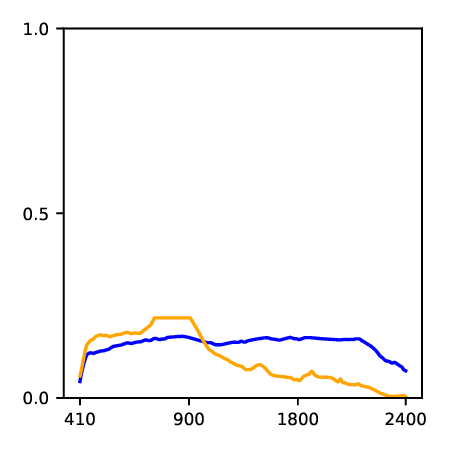}
    \hspace{-0.35cm}
    \includegraphics[width=0.128\linewidth]{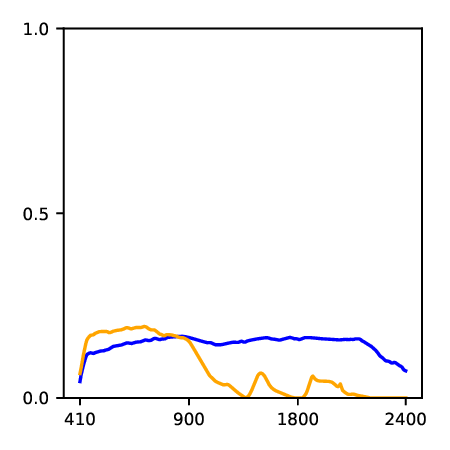}
    \hspace{-0.35cm}
    \includegraphics[width=0.128\linewidth]
    {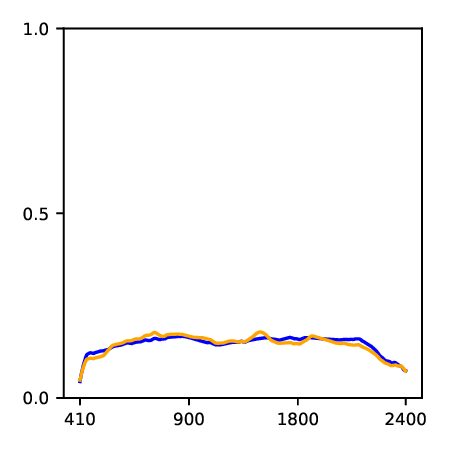}
\\
 \rotatebox{90}{\scriptsize{\hspace{0.9cm}Road}}
    \includegraphics[width=0.128\linewidth]{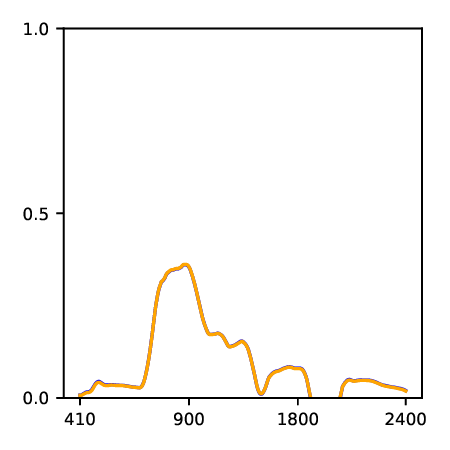}
    \hspace{-0.35cm}
    \includegraphics[width=0.128\linewidth]{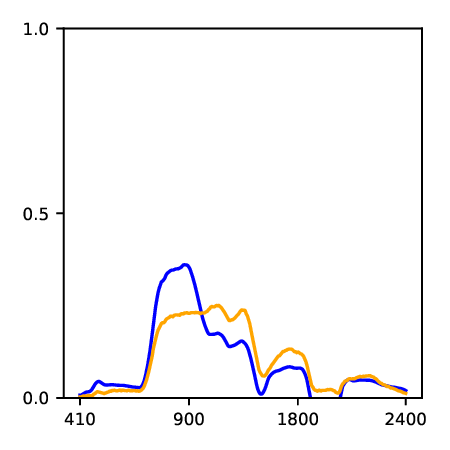}
    \hspace{-0.35cm}
    \includegraphics[width=0.128\linewidth]{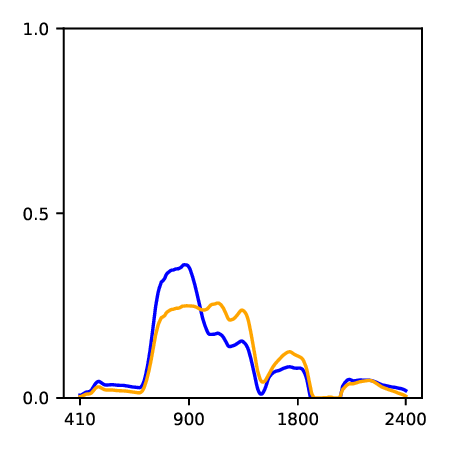}
    \hspace{-0.35cm}
    \includegraphics[width=0.128\linewidth]{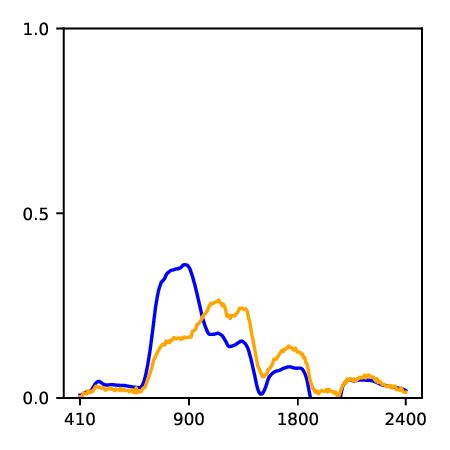}
    \hspace{-0.35cm}
    \includegraphics[width=0.128\linewidth]{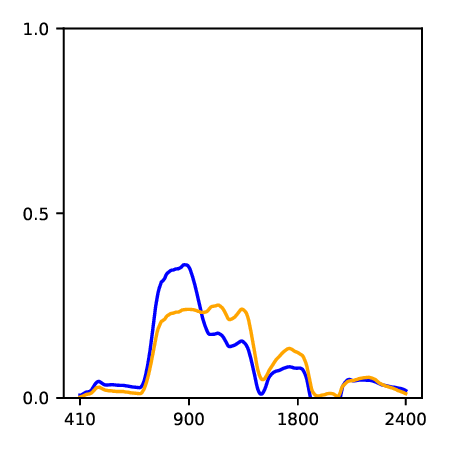}
    \hspace{-0.35cm}
    \includegraphics[width=0.128\linewidth]{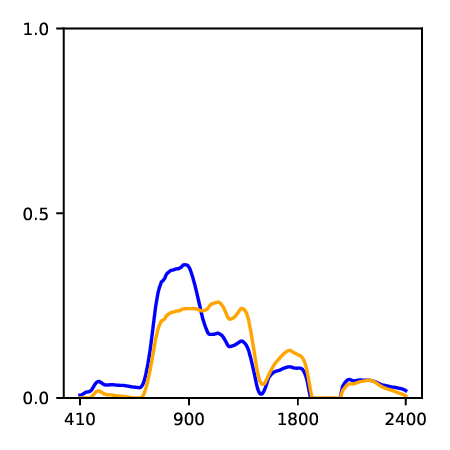}
    \hspace{-0.35cm}
    \includegraphics[width=0.128\linewidth]{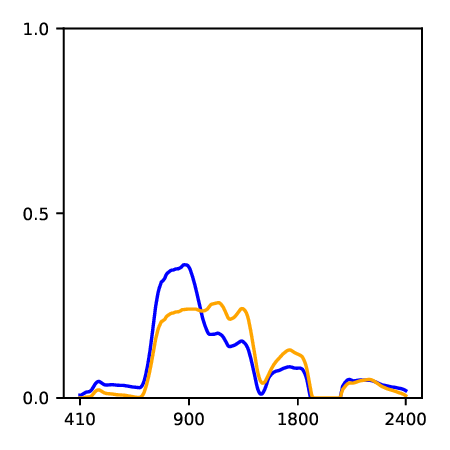}
    \hspace{-0.35cm}
    \includegraphics[width=0.128\linewidth]
    {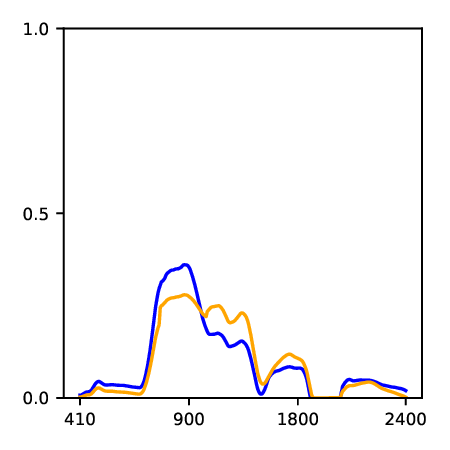}
\\
 \rotatebox{90}{\scriptsize{\hspace{1cm}Tree}}
    \includegraphics[width=0.128\linewidth]{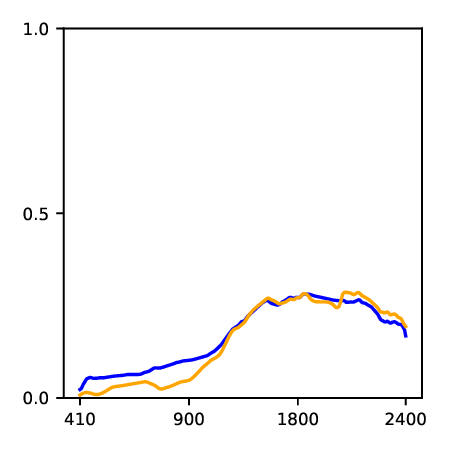}
    \hspace{-0.35cm}
    \includegraphics[width=0.128\linewidth]{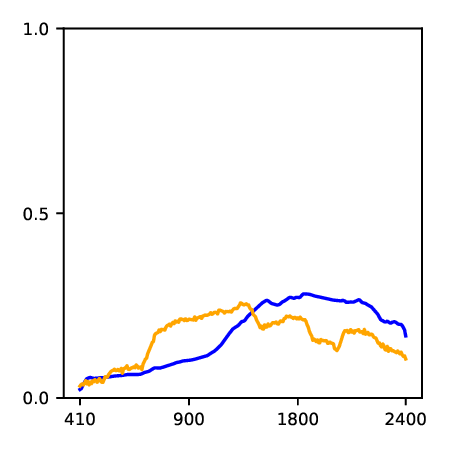}
    \hspace{-0.35cm}
    \includegraphics[width=0.128\linewidth]{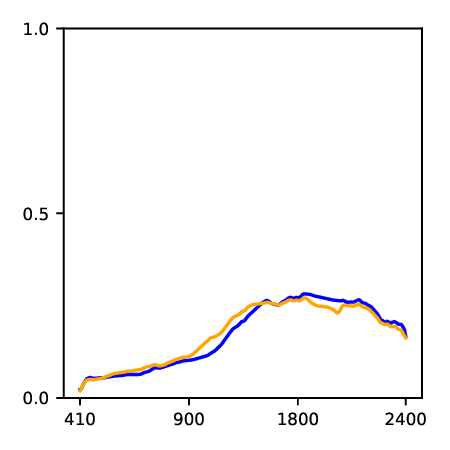}    \hspace{-0.35cm}
    \includegraphics[width=0.128\linewidth]{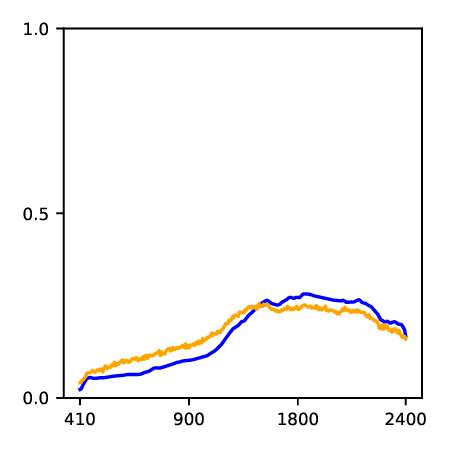}
    \hspace{-0.35cm}
    \includegraphics[width=0.128\linewidth]{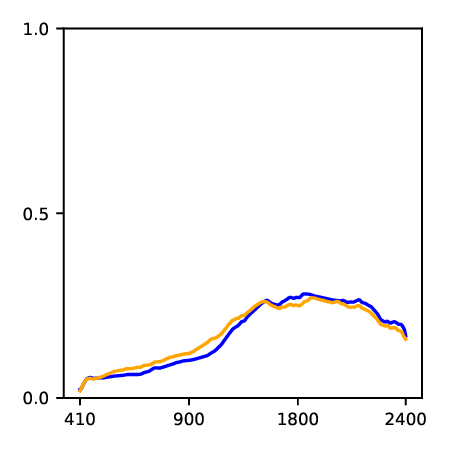}
    \hspace{-0.35cm}
    \includegraphics[width=0.128\linewidth]{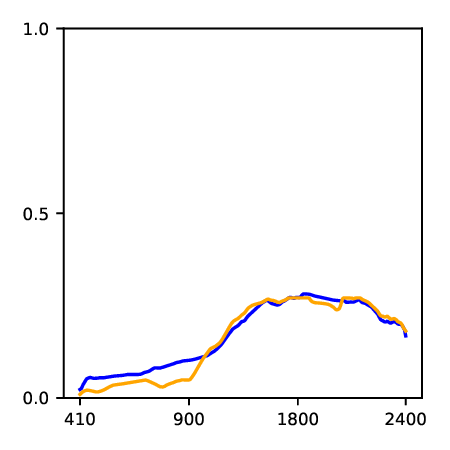}
    \hspace{-0.35cm}
    \includegraphics[width=0.128\linewidth]{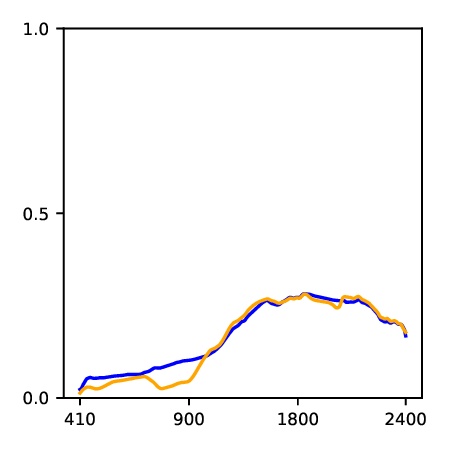}
    \hspace{-0.35cm}
    \includegraphics[width=0.128\linewidth]
    {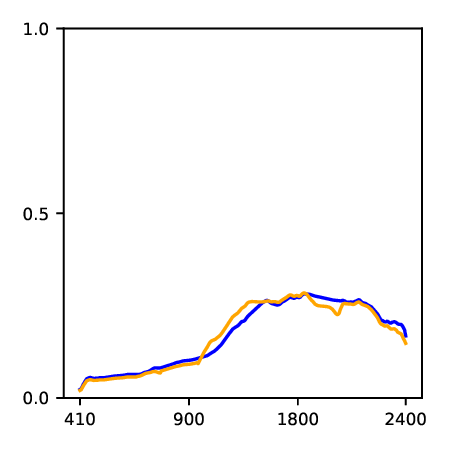}
\\
 \rotatebox{90}{\scriptsize{\hspace{0.9cm}Water}}
    \includegraphics[width=0.128\linewidth]{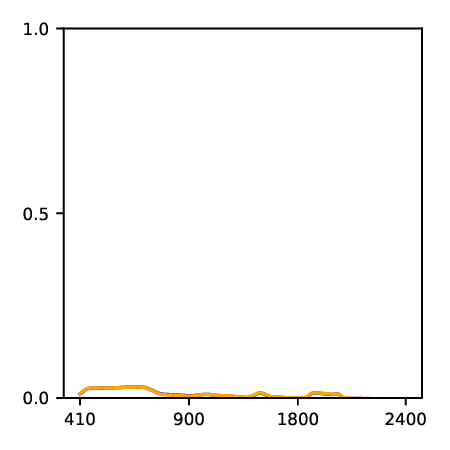}
    \hspace{-0.35cm}
    \includegraphics[width=0.128\linewidth]{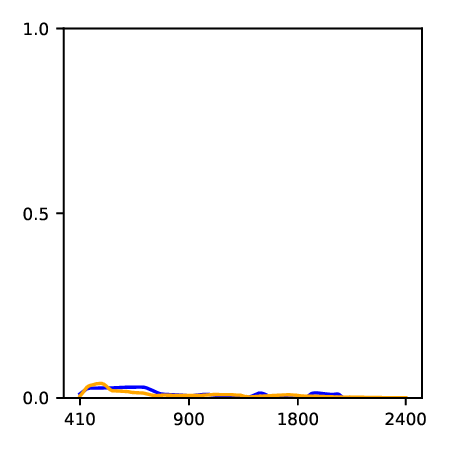}    \hspace{-0.35cm}
    \includegraphics[width=0.128\linewidth]{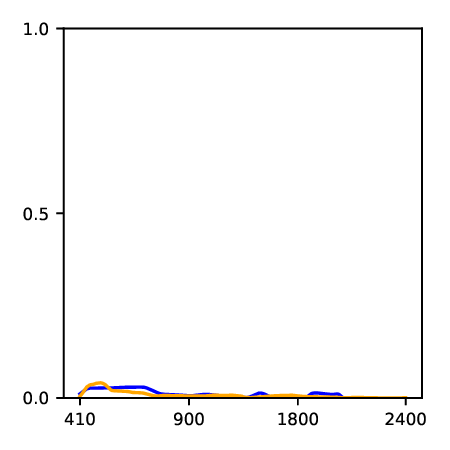}
    \hspace{-0.35cm}
    \includegraphics[width=0.128\linewidth]{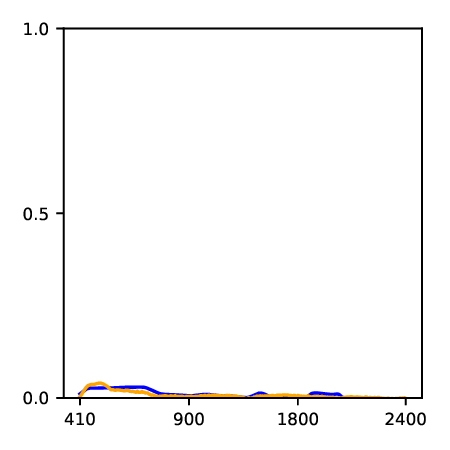}
    \hspace{-0.35cm}
    \includegraphics[width=0.128\linewidth]{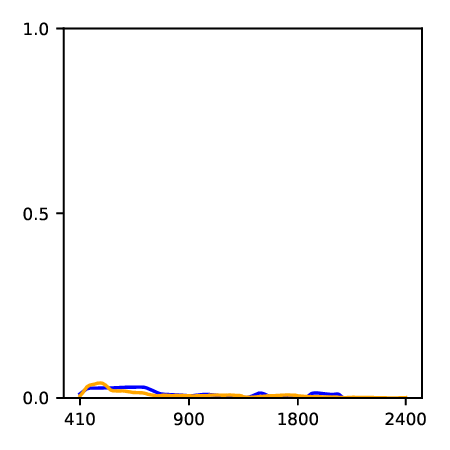}
    \hspace{-0.35cm}
    \includegraphics[width=0.128\linewidth]{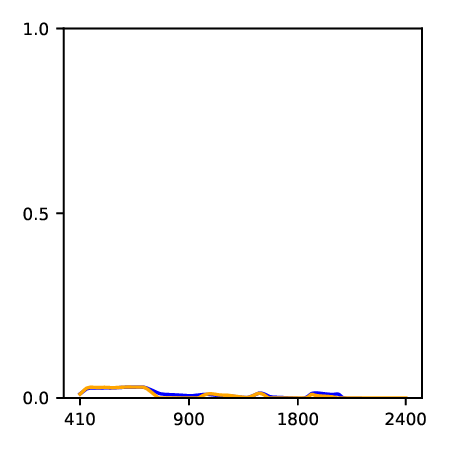}
    \hspace{-0.35cm}
    \includegraphics[width=0.128\linewidth]{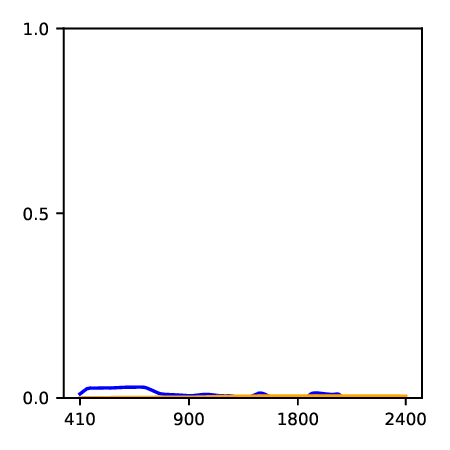}
    \hspace{-0.35cm}
    \includegraphics[width=0.128\linewidth]
    {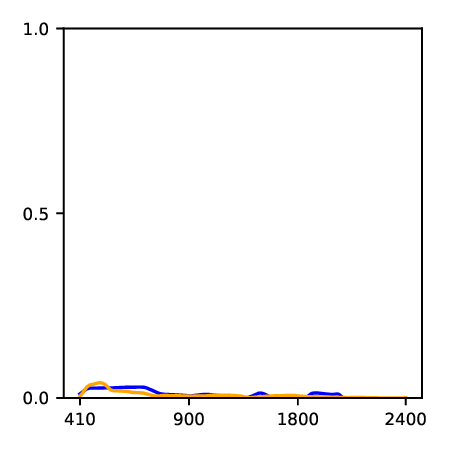}
    \begin{flushleft}
    \scriptsize{\hspace{0.5cm}FCLSU~\cite{heinz1999fully, nascimento2005vertex} \hspace{0.6cm} CNNAEU~\cite{palsson2020convolutional} \hspace{0.5cm} DeepTrans~\cite{9848995} \hspace{0.6cm} EDAA~\cite{zouaoui2023entropic} \hspace{0.8cm} EndNet ~\cite{ozkan2018endnet}\hspace{0.8cm} MiSiCNet~\cite{rasti2022misicnet} \hspace{0.6cm} Hyperweak~\cite{shen2022toward} \hspace{0.6cm} Ours}
    \end{flushleft}
    \caption{Visual comparison of endmember spectra obtained by different unmixing techniques. The blue curves correspond to the ground‐truth spectra, while the orange curves denote the spectra estimated by each method.}
    \label{fig:enter-label}
\end{figure*}


Tables~\ref{tab:apex_rmse} and~\ref{tab:apex_sad} present the performance comparison on the APEX dataset using RMSE and SAD metrics. Our method excels precisely where weak signals dominate (e.g., Road with small fractional cover and Water with low reflectance). It attains the lowest mean RMSE = 0.0460  and the lowest mean SAD = 0.0740. Per-endmember, we achieve the best RMSE on Tree (0.0331), Road (0.0167), and Water (0.0841) and are second on Roof; for SAD we are best across all four endmembers. These gains reflect the model’s weak-signal enhancement and multi-wavelet feature extraction: compared with the next best method, RMSE drops by 31\% on Road and 6.8\% on Water, while preserving spectral shape (best SAD) for every class.

The Apex visualization Fig.~\ref{fig:apex_abundance_maps} and Fig.~\ref{fig:enter-label} shown, , our proposed method demonstrates superior robustness, particularly in scenarios dominated by weak signals such as Roof (small fractional cover) and Water (low reflectance). We observe that, on the Roof endmember, the abundance map estimated by our approach exhibits the closest alignment to the ground truth. Similarly, on the Water endmember, representing a weak spectral response, our method achieves more competitive accuracy compared with existing baselines. This significant improvement in overall network performance can be primarily attributed to the collaborative integration of the MambaVision and Transformer architectures, reinforced by the proposed weak-signal attention mechanism. Through this synergy, the framework effectively captures both global contextual dependencies and fine-grained local details, leading to substantial gains in endmember discrimination and abundance estimation accuracy under challenging conditions.

\begin{table}[htbp]
\centering
\scriptsize
\caption{Root Mean Square Error (RMSE) comparison on the APEX dataset. RMSE is computed for each endmember and averaged across all four. Lower is better. Best results are in \textcolor{red}{\textbf{bold red}}.}
\begin{tabular}{@{}lccccc@{}}
\toprule
\textbf{Method} & Roof & Tree & Road & Water & \textbf{RMSE\_Mean} \\
\midrule
FCLSU~\cite{heinz1999fully, nascimento2005vertex}        & 0.0942 & 0.0554 & 0.1526 & 0.1746 & 0.1192 \\
Hyperweak~\cite{shen2022toward} & 0.2131 & 0.1743 & 0.2109 & 0.4528 & 0.2628 \\
CNNAEU~\cite{palsson2020convolutional}       & 0.0588 & 0.0385 & 0.0769 & 0.1233 & 0.0743 \\
DeepTrans~\cite{9848995}    & 0.1200 & 0.0993 & 0.1776 & 0.0902 & 0.1218 \\
EDAA~\cite{zouaoui2023entropic}         & 0.0947 & 0.0701 & 0.0311 & 0.1702 & 0.0915 \\
EndNet ~\cite{ozkan2018endnet}      & \textcolor{red}{\textbf{0.0337}} & 0.0346 & 0.1071 & 0.1058 & 0.0703 \\
MiSiCNet~\cite{rasti2022misicnet}     & 0.0771 & 0.0385 & 0.0242 & 0.2091 & 0.0872 \\
\rowcolor{gray!10}
Ours         & 0.0502 & \textcolor{red}{\textbf{0.0331}} & \textcolor{red}{\textbf{0.0167}} & \textcolor{red}{\textbf{0.0841}} & \textcolor{red}{\textbf{0.0460}}\\
\bottomrule
\end{tabular}
\label{tab:apex_rmse}
\end{table}

\begin{table}[htbp]
\centering
\scriptsize
\caption{Spectral Angle Distance (SAD) comparison on the APEX dataset. Lower is better. Best results are in \textcolor{red}{\textbf{bold red}}.}
\begin{tabular}{@{}lccccc@{}}
\toprule
\textbf{Method} & Roof & Tree & Road & Water & \textbf{SAD\_Mean}\\
\midrule
FCLSU~\cite{heinz1999fully, nascimento2005vertex}        & 0.6900 & 0.2494 & 0.1471 & 0.5058 & 0.3981 \\
Hyperweak~\cite{shen2022toward} & 0.0964 & 0.5448 & 0.1700 & 0.0964 & 0.6155 \\
CNNAEU~\cite{palsson2020convolutional}       & 0.2649 & 0.1724 & 0.4133 & 0.0853 & 0.2340 \\
DeepTrans~\cite{9848995}    & 0.0836 & 0.1295 & 0.0903 & 0.0434 & 0.0867 \\
EDAA~\cite{zouaoui2023entropic}         & 0.6943 & 0.3178 & 0.1633 & 0.1365 & 0.3280 \\
EndNet ~\cite{ozkan2018endnet}      & 0.2296 & 0.1550 & 0.0893 & 0.0839 & 0.1394 \\
MiSiCNet~\cite{rasti2022misicnet}     & 0.5475 & 0.1725 & 0.1269 & 0.1757 & 0.2556 \\
\rowcolor{gray!10}
Ours         & \textcolor{red}{\textbf{0.0630}} & \textcolor{red}{\textbf{0.1031}} & \textcolor{red}{\textbf{0.0875}} & \textcolor{red}{\textbf{0.0424}} & \textcolor{red}{\textbf{0.0740}} \\
\bottomrule
\end{tabular}
\label{tab:apex_sad}
\end{table}

\subsection{Ablation Studies}

Ablation Study of Encoder Variants
Table~\ref{tab:ablation1} presents the progressive effects of incorporating wavelet branches into the baseline convolutional model. The Conv-only setting yields the highest reconstruction errors, particularly on weak-signal metrics (\(w\)SAD, \(w\)RMSE), reflecting a bias toward dominant spectral responses. Adding the Haar branch improves spectral fidelity on most datasets, although its performance varies in Samson, indicating dataset-dependent sensitivity. When both Haar and Symlet-3 branches are integrated, the model consistently achieves the lowest errors, with weak-signal RMSE reduced by over 85\% on S1 and more than 90\% on Apex. These findings demonstrate that Haar and Symlet-3 provide complementary representations; Haar enhances spectral discontinuities while Symlet-3 preserves smooth variations, thereby enabling robust preservation of weak spectral signatures under complex mixing conditions.

\begin{table}[]
\centering
\scriptsize
\label{tab:ablation1}
\caption{Ablation study of multi-resolution wavelet modules. Weak signal SAD (wSAD )and Weak signal RMSE (wRMSE) denote weak-signal errors.}
\begin{tabular}{|c|l|l|l|l|l|l|l|}
\hline
\multicolumn{1}{|l|}{}  & Conv & Haar & Symlet-3 & \(a\)SAD   &\(a\)RMSE  & \(w\)SAD   & \(w\)RMSE  \\ \hline
\multirow{3}{*}{S1}     & \checkmark    &      &       & 0.2675 & 0.2875 & 0.1792 & 0.2104 \\ \cline{2-8} 
                        & \checkmark    & \checkmark    &       & 0.0484 & 0.0638 & 0.1024 & 0.0874 \\ \cline{2-8} 
                        & \checkmark    & \checkmark    & \checkmark     & 0.0271 & 0.0181 & 0.0262 & 0.0152 \\ \hline
\multirow{3}{*}{Samson} & \checkmark    &      &       & 0.1025 & 0.0785 & 0.0957 & 0.1471 \\ \cline{2-8} 
                        & \checkmark    & \checkmark    &       & 0.0914 & 0.1470 & 0.1705 & 0.1142 \\ \cline{2-8} 
                        & \checkmark    & \checkmark    & \checkmark     & 0.0378 & 0.0736 & 0.0436 & 0.0153 \\ \hline
\multirow{3}{*}{Apex}   & \checkmark    &      &       & 0.2701 & 0.1987 & 0.1477 & 0.2441 \\ \cline{2-8} 
                        & \checkmark    & \checkmark    &       & 0.0994 & 0.1107 & 0.1906 & 0.1439 \\ \cline{2-8} 
                        & \checkmark    & \checkmark    & \checkmark     & 0.0747 & 0.0302 & 0.0424 & 0.0208 \\ \hline

\end{tabular}
\end{table}
To further understand the contribution of each architectural component, we conduct a comprehensive ablation study on top of the fixed multi-resolution wavelet encoder (Conv-only, Conv+Haar, and Conv+Haar+Symlet-3). All results reported in Table~\ref{tab:ablation_mti} use the strongest encoder variant (Conv+Haar+Symlet-3), ensuring that the backbone differences are isolated from the effects of spectral–spatial feature extraction.

The results reveal several important observations. First, using any single branch already brings notable improvements over the Mamba-only baseline, indicating that both the Transformer backbone (T) and the Weak Signal Attention branch (IAT) contribute significantly to stabilizing the reconstruction of weak spectral components. Among single branches, IAT provides the most consistent gains in both RMSE and SAD across all three datasets, confirming its ability to enhance weak-signal detection through localized Weak Signal Attention modulation.

Second, combining Mamba with either T or IAT (V4 and V5) yields further performance gains, demonstrating that state-space modeling is complementary to both global patch-level attention and weak-signal attention refinement. Notably, the T+IAT configuration (V6) achieves the best performance among all two-branch variants and closely approaches the full model, showing that the interaction between global self-patch attention and weak-signal refinement is particularly beneficial.

Finally, enabling all three components simultaneously (V7) achieves the lowest RMSE and SAD on all datasets, especially on S1 and Samson where weak endmembers are more challenging. This confirms that Mamba’s global sequence modeling, the Transformer’s multi-scale patch attention, and the weak-signal Weak Signal Attention mechanism each address different aspects of the unmixing problem, and their integration leads to the strongest overall robustness against weak-signal collapse and spectral variability.
\begin{table*}[t]
\centering
\caption{Ablation of Mamba state-space (M), Multihead Self-Patch Attention (T), and Weak Signal Attention (WAT) on three datasets. 
For each variant, we report the average RMSE and SAD on S1, Samson, and APEX.}
\label{tab:ablation_mti}
\setlength{\tabcolsep}{6pt}
\begin{tabular}{c l c c c cc cc cc}
\toprule
\multirow{2}{*}{ID} & \multirow{2}{*}{Variant} & \multirow{2}{*}{M} & \multirow{2}{*}{T} & \multirow{2}{*}{IAT}
& \multicolumn{2}{c}{S1} & \multicolumn{2}{c}{Samson} & \multicolumn{2}{c}{APEX} \\
\cmidrule(lr){6-7} \cmidrule(lr){8-9} \cmidrule(lr){10-11}
& & & & & RMSE$\downarrow$ & SAD$\downarrow$
& RMSE$\downarrow$ & SAD$\downarrow$
& RMSE$\downarrow$ & SAD$\downarrow$ \\
\midrule
V1 & M only              & \cmark & \xmark & \xmark & 0.1742 & 0.2028 & 0.2141 & 0.2702 & 0.2204 & 0.3052 \\
V2 & T only              & \xmark & \cmark & \xmark & 0.0270 & 0.0252 & 0.0784 & 0.1025 & 0.1066 & 0.0871 \\
V3 & WAT only            & \xmark & \xmark & \cmark & 0.0281 & 0.0302 & 0.0544 & 0.0925 & 0.0981 & 0.0891 \\
V4 & M + T               & \cmark & \cmark & \xmark & 0.0762 & 0.1824 & 0.0612 & 0.0670 & 0.0745 & 0.0810 \\
V5 & M + WAT             & \cmark & \xmark & \cmark & 0.0527 & 0.0225 & 0.0478 & 0.0589 & 0.0621 & 0.0772 \\
V6 & T + WAT             & \xmark & \cmark & \cmark & 0.0201 & 0.0638 & 0.0410 & 0.0458 & 0.0535 & 0.0758 \\
V7 & M + T + WAT (Full)  & \cmark & \cmark & \cmark & 0.0131 & 0.0221 & 0.0368 & 0.0378 & 0.0460 & 0.0740 \\
\bottomrule
\end{tabular}
\end{table*}


\section{Conclusion and Future Work}
This study examined weak-signal hyperspectral unmixing from both a theoretical and algorithmic perspective. By revisiting the linear mixing model under low-energy endmembers, we showed that weak-signal regimes amplify the nonlinear residual term and effectively turn abundance recovery into an ill-posed inverse problem. This explains why conventional encoders suppress low-magnitude spectra and why abundance estimates become unstable when materials with weak signals coexist with dominant components or sensor noise. Building on this formulation, WS-Net integrates a multi-resolution wavelet encoder with a hybrid Mamba–Weak Signal Attention backbone to stabilise the inversion. The wavelet pathway preserves high-frequency structures associated with materials that exhibit weak spectral responses, while the state-space and Weak Signal Attention components collectively retain subtle spectral cues that typically collapse under noise or similarity-driven attention. The sparsity-aware decoder, supported by a KL-based spectral-shape regulariser, further constrains the solution space and mitigates ambiguities intrinsic to weak-signal mixtures. Across simulated and two real datasets (Samson and Apex), WS-Net consistently improves abundance estimation and endmember fidelity relative to classical and recent deep unmixing models. The gains are most pronounced for materials with weak signals, and the method remains robust across a wide range of SNR conditions, reinforcing the value of treating weak-signal unmixing as a structurally ill-posed problem requiring explicit regularisation.

WS-Net offers a flexible foundation for weak-signal modelling in remote sensing, and several directions remain open. A deeper theoretical characterisation of how nonlinear residuals interact with low-energy endmembers would help clarify the limits of linear approximations and guide the design of more principled regularisers. Scaling the framework to large airborne and satellite scenes, where endmember variability, illumination geometry, and atmospheric distortions are more pronounced, would allow a fuller assessment of weak-signal behaviour in operational settings. Another promising direction is cross-sensor and cross-scene adaptation, which could reduce the reliance on dataset-specific calibration and make weak-signal unmixing more robust across heterogeneous platforms. Advancing along these lines would strengthen the connection between physical modelling and learned representations, and broaden the reliability and reach of weak-signal unmixing in real-world applications.

\bibliographystyle{IEEEtran}
\bibliography{ref}


\begin{IEEEbiography}[{\includegraphics[width=1in,height=1.25in,clip,keepaspectratio,angle=-90]{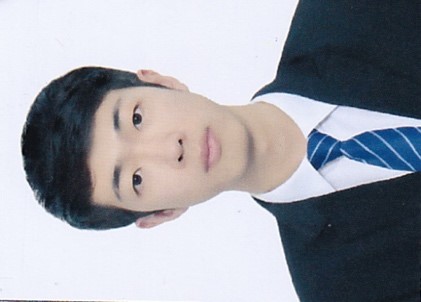}}]{Zekun Long}
received the Master of Information Technology degree in Data Analytics from Griffith University, Brisbane, QLD, Australia, in 2022. He is currently pursuing the Ph.D. degree with the School of Information and Communication Technology at Griffith University. His research interests include hyperspectral image processing, spectral unmixing, and remote sensing analytics. His recent work focuses on weak-signal hyperspectral modeling, deep learning architectures, and diffusion-based spectral priors for quantitative environmental and food-safety analysis.
\end{IEEEbiography}

\begin{IEEEbiography}[{\includegraphics[width=1in,height=1.25in,clip,keepaspectratio]{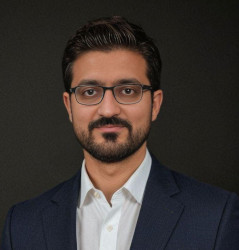}}]{Ali Zia}
is a Research Fellow in the School of Computing, Engineering and Mathematical Sciences at La Trobe University. He received the B.Sc. (Hons.) degree in computer science from Punjab University (Pakistan), the Master of Computing degree from the Australian National University, and the Ph.D. degree from Griffith University, Australia. He completed a Postdoctoral Fellowship at the Australian National University with an affiliation at CSIRO Australia. His research focuses on higher-order representation learning, hyperspectral imaging, and computer vision, with an emphasis on topology-aware and weakly supervised methods for robust perception. He has published more than fifty peer-reviewed papers in leading venues, including IEEE Transactions on Image Processing, Artificial Intelligence Review, and ICCV.
\end{IEEEbiography}

\begin{IEEEbiography}[{\includegraphics[width=1in,height=1.25in,clip,keepaspectratio]{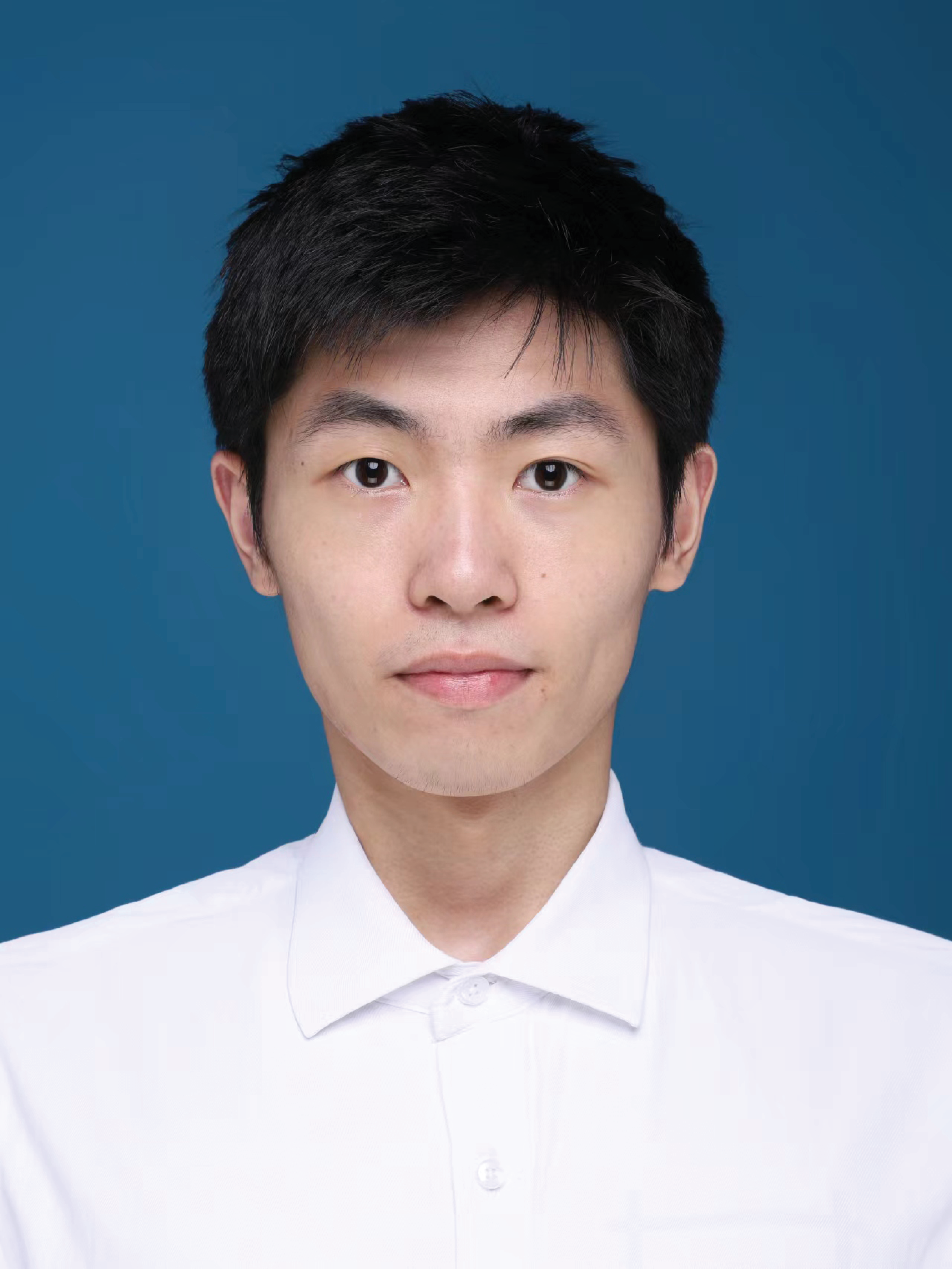}}]{Guanyiman Fu}
received the B.E. degree from the College of Life Sciences, Northwest Agriculture and Forestry University, Xianyang, China, in 2017. He is currently pursuing the Ph.D. degree in computer science and technology at Nanjing University of Science and Technology, Nanjing, China, and is a visiting student at the School of Information and Communication Technology, Griffith University, Brisbane, Australia. His research interests include hyperspectral image restoration and hyperspectral image processing.
\end{IEEEbiography}

\begin{IEEEbiography}[{\includegraphics[width=1in,height=1.25in,clip,keepaspectratio]{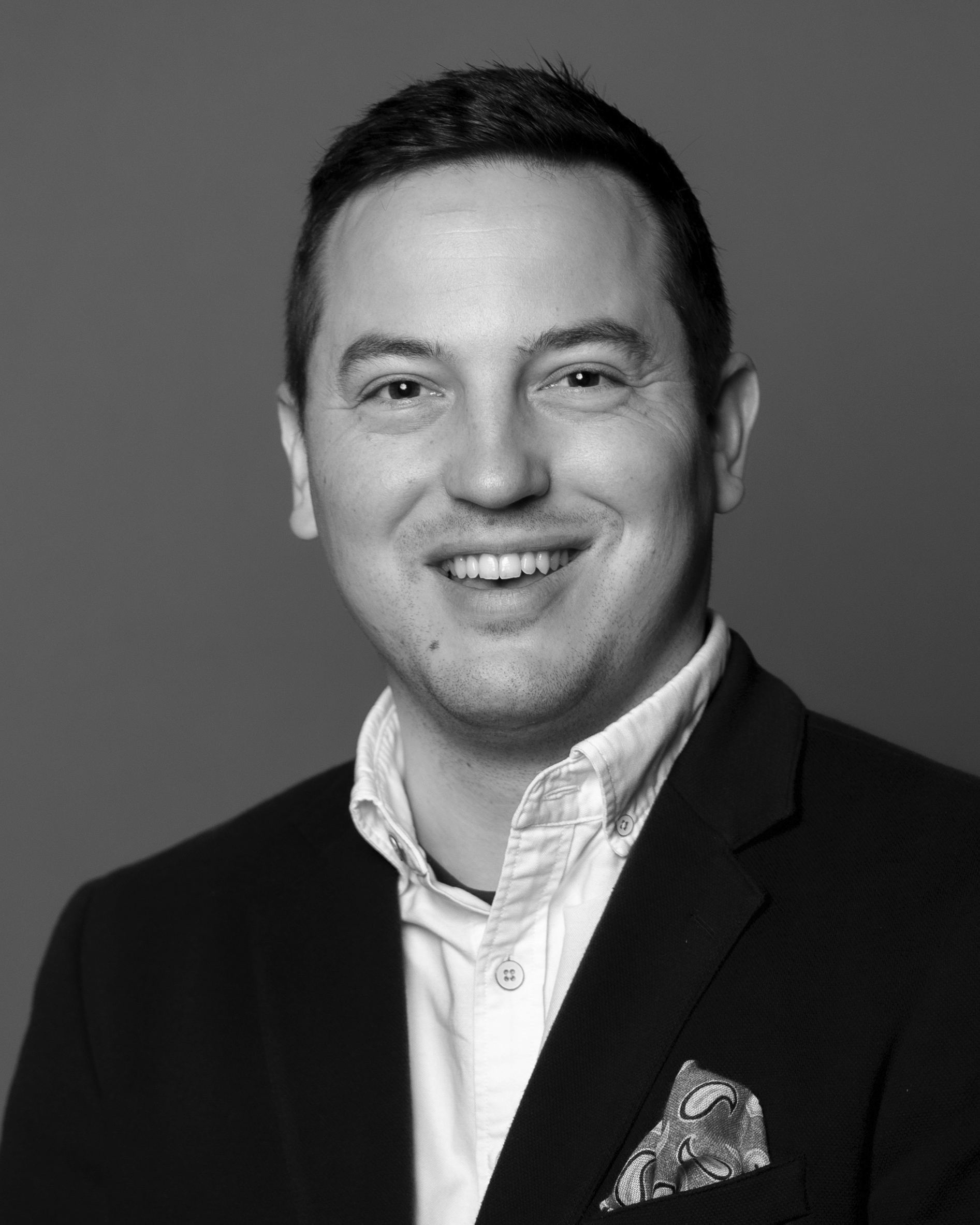}}]{Vivien Rolland}
is a Senior Research Scientist and the leader of the Digital Crops Team at CSIRO, Agriculture and Food. His work focuses on using machine learning and computer vision to accelerate innovation in the agrifood sector, including the development of digital tools that support next-generation crop breeding. His research spans multimodal sensing, plant phenomics, and large-scale environmental analytics.
\end{IEEEbiography}

\begin{IEEEbiography}[{\includegraphics[width=1in,height=1.25in,clip,keepaspectratio]{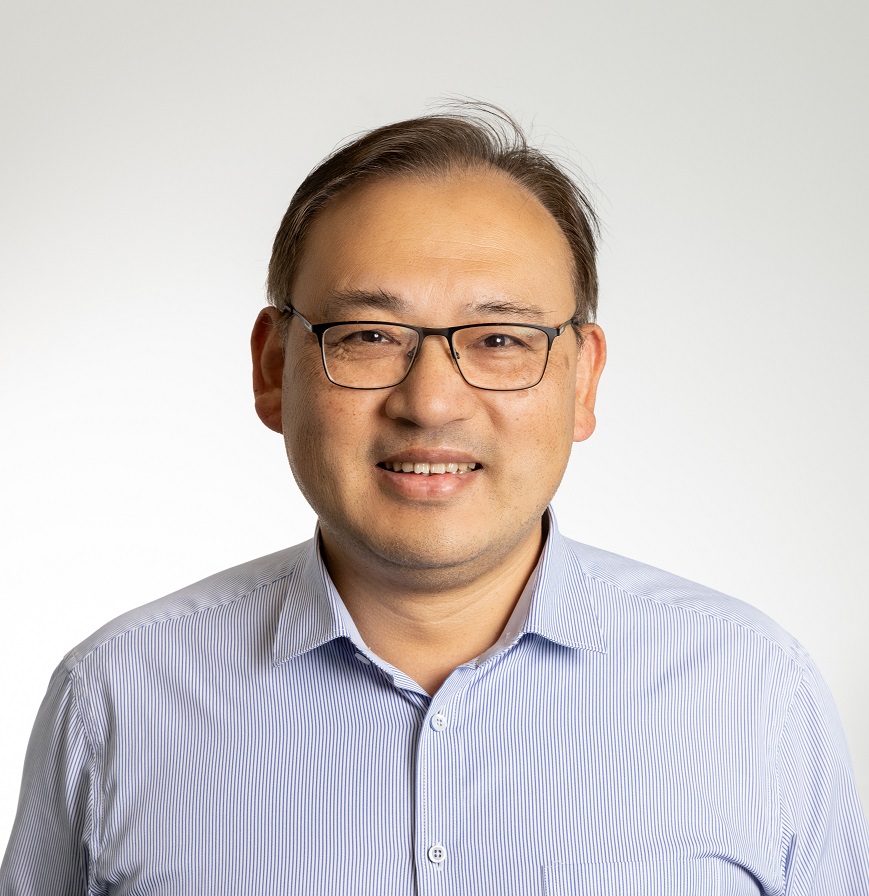}}]{Jun Zhou (Fellow, IEEE)}
received the B.S. degree in computer science and the B.E. degree in international business from Nanjing University of Science and Technology, Nanjing, China, in 1996 and 1998, respectively. He received the M.S. degree in computer science from Concordia University, Montreal, Canada, in 2002, and the Ph.D. degree from the University of Alberta, Edmonton, Canada, in 2006.

He is currently a Professor with the School of Information and Communication Technology, Griffith University, Nathan, Australia. Previously, he was a Research Fellow with the Research School of Computer Science at the Australian National University and a Researcher with the Canberra Research Laboratory, NICTA, Australia. His research interests include pattern recognition, computer vision, and hyperspectral image processing with applications to remote sensing and environmental informatics. He is an Associate Editor of \textsc{IEEE Transactions on Geoscience and Remote Sensing} and \textsc{Pattern Recognition}, and an IEEE Fellow.
\end{IEEEbiography}
\end{document}